\documentclass[lettersize,journal]{IEEEtran}

\usepackage{import}
\usepackage{albi_pack}
\usepackage{ulem}

\makenomenclature
\newtheorem {property}{Property}
\newtheorem {assumption}{Assumption}
\begin{document}


\title{Probabilistic Safety Regions Via Finite Families of Scalable Classifiers}



\author{Alberto~Carlevaro,~\IEEEmembership{Student Member,~IEEE},
        Teodoro~Alamo,~\IEEEmembership{Member,~IEEE}, 
        ~Fabrizio~Dabbene,~\IEEEmembership{ Senior Member,~IEEE} and 
        Maurizio~Mongelli,~\IEEEmembership{Member,~IEEE} 
        

\IEEEcompsocitemizethanks{\IEEEcompsocthanksitem A.C., F.D., and M.M. are with Cnr-Istituto di Elettronica e di Ingegneria dell’Informazione e delle Telecomunicazioni, 00185 Rome, Italy. A.C. is with University of Genoa, Department of Electrical, Electronics and Telecommunications Engineering and Naval Architecture (DITEN), 16145 Genoa, Italy. Teodoro Alamo is with the Departamento de Ingeniería de Sistemas y Automática, Universidad de Sevilla, Escuela Superior de Ingenieros, 41020 Sevilla, Spain.
\IEEEcompsocthanksitem Corresponding author: Alberto Carlevaro, alberto.carlevaro@ieiit.cnr.it, alberto.carlevaro@edu.unige.it.\protect
\IEEEcompsocthanksitem E-mail: talamo@us.es, fabrizio.dabbene@ieiit.cnr.it, maurizio.mongelli@ieiit.cnr.it.}


\thanks{This work has been submitted to the IEEE for possible publication. Copyright may be transferred without notice, after which this version may no longer be accessible.
}}


\markboth{IEEE Transactions on Neural Networks and Learning Systems,~Vol.~XX, No.~XX, Month~Year}%
{Shell \MakeLowercase{\textit{et al.}}: Bare Demo of IEEEtran.cls for Computer Society Journals}


\IEEEtitleabstractindextext{%
\begin{abstract}
Supervised classification recognizes patterns in the
data to separate classes of behaviours. Canonical solutions
contain misclassification errors that are intrinsic to the numerical approximating nature of
machine learning. The data analyst may minimize the classification error on a class at the expense of increasing the
error of the other classes. The error control of such a design phase is
often done in a heuristic manner. In this context, it is key to develop theoretical foundations capable of providing probabilistic certifications to the obtained classifiers. In this perspective, we introduce the concept of probabilistic safety region to describe a subset of the input space in which the number of misclassified instances is probabilistically controlled.
The notion of scalable classifiers is then exploited to link the
tuning of machine learning with error control. Several tests
corroborate the approach. They are provided through synthetic
data in order to highlight all the steps involved, as well as
through a smart mobility application.
\end{abstract}


}

                
\maketitle

\IEEEdisplaynontitleabstractindextext

\IEEEpeerreviewmaketitle

                


\section{Introduction and Problem Formulation}\label{sec:introduction}

\IEEEPARstart{S}{afety-critical} assessment of machine learning (ML) is currently one of the main issues in trustworthy artificial intelligence \cite{Amodei,9074237}. The scope is to understand under which conditions autonomous operation may lead to hazards, in order to reduce to the minimum the risk of operating with detrimental effects to the human or the environment. Such an assessment is mandatory in several application domains, such as avionics \cite{doi:10.2514/6.2020-1521}, finance\cite{RePEc:ofr:wpaper:15-19}, healthcare\cite{zou2023review}, smart mobility \cite{safeautonomousdriving}, cybersecurity\cite{Jorgensen_2023} as well as with autonomous systems \cite{Amodei,BoLi,9074237}. Informally speaking, the safety assurance of ML consists of building \textit{guardrails} around the autonomous decision in front of uncertainty \cite{BruceGuardrail}. This can be achieved with a combination of rigorous verification \cite{LUCAS20084591}, design validation \cite{8595071} or standardized safety analysis\footnote{\url{https://standards.ieee.org/practices/mobility/standards/}.}. However, those approaches look at ML from the outside, by analyzing risks at system level. Intrinsically in ML, on the other hand, there is the possibility to link the search of classification boundaries with error control. More specifically, the error control may be built around the class of behavior it is desired to protect, e.g., collision avoidance, disease status, cyber attack in progress. This may change the theoretical approach to algorithm design considerably. In this respect, the guardrails are built here on the basis of \textit{order statistics} \cite{David2011}, by incorporating them as probabilistic constraints of the model. The notion of \textit{probabilistic safety region} is firstly outlined. The region defines the subset of data samples under which the error is constrained. 
Second, a specific class of classifiers, named \textit{scalable classifiers} is introduced as a valuable tool to construct probabilistic safety regions.
In particular, the optimal shape of the region may be obtained by setting the value of the parameters within the set of possible scalable classifiers.

Namely, scalable classifiers share the property of having a scaling parameter that can be adjusted to control both the classification boundary and the inherent error. Suitable classifiers are in this respect, for instance, support vector machines, support vector data description, logistic regression; but the class is very rich and several different and new scalable classifiers may be devised.
Probabilistic scaling \cite{MAMMARELLA2022110108} drives the validation of parameters setting. Based on this scaling procedure, the proposed design methodology is capable of probabilistically guaranteeing a given error level valid for the largest possible region.

\subsection{Contribution}\label{sec:contribution}
More specifically, the contribution of this research is twofold. First, the family of \textit{Scalable Classifiers} (SCs) is studied as a new group of classifiers that share the property of being scalable, that is, of being controllable by a single scalar parameter (Section \ref{subseq:scalable_class}). Then, the definition of \textit{Probabilistic Safety Region} (PSR) models the idea of giving probabilistic guarantees on the prediction of a classification (but in general also a regression) problem (Section \ref{subseq:PSR}). The link between these two concepts is provided by \textit{probalistic scaling}, a state-of-the-art technique for providing probabilistic bounds based on the field of order statistics \cite{Alamo2018}. Moreover, we take into account the variability of the classifier due to the hyperparameter selection. It is known that the choice of the hyperparameters can affect dramatically the result in the prediction of the model. Usually, their optimal setting is a hard task. To cope with this problem, we apply our probabilistic safety framework to classifiers obtained by \textit{finite families} of hyperparameters, selecting among this set of confident classifiers the one which optimizes a certain statistical index, for example minimizing the false positive rate of the classification (Section \ref{sec:families}).\\
The result is a totally new framework in statistical learning that shares the requirements of AI trustworthiness, i.e., it is reliable, safe and robust.
\subsection{Related works}\label{sec:sota}
The concept of \textit{robust} Machine Learning has several subtle meanings. Initially, overfitting was the main issue involved in robustness \cite{anderson2004model}. More recently, robustness passes through the capacity of the model to counter adversarial attacks \cite{9149002}, to handle data-privacy problems \cite{DBLP1}, to have good generalisation properties \cite{DBLP2}, and many other theoretical and practical challenges \cite{DBLP3} facing the scientific community.
Today, robustness is inseparable from safety, and the reason for this is entirely agreeable: algorithms that cannot handle data fluctuations or fail to provide sufficient levels of reliability can lead to risky situations for users. Safety and robustness are, therefore, fundamental requirements that cannot be separated.
\\
In relation with the concept of safety \cite{7888195}, the idea of \textit{Safety Region} (SR) \cite{doi:10.1177/1687814016668140} deals with the identification of the regions of the input space that lead to predictions with the same guaranteed level of confidence. Currently, there are different methodologies addressing this topic:
\textit{Conformal Prediction} (CP) \cite{JMLR:v9:shafer08a}, which deals with the discovery of the confidence interval of each prediction, thus giving a clear indication of the quality of the prediction.
\textit{Error Quantification} \cite{MirasierraUQ} provides a methodology to quantify uncertainty with probabilistic maximization. 
\textit{Bayesian learning contexts} \cite{10.5555/2207809}, which provide probabilistic guarantees by exploiting the degree of belief about the output.
\textit{Selective Classification}, \cite{10.2307/20445230} in which the model can abstain from making a prediction and the goal is to minimize incorrect predictions while also minimizing abstentions.
\textit{Covariate shift} \cite{10.5555/1462129}, where the goal is to learn a classifier based on samples
    from training and test that have different marginal distributions but share the same conditional labelling functions for all points.
CP is one of the most recognised methods for confidence calculation. In order to meet the probabilistic guarantees, CP might assign more than a single class to a given point of the input space. The method proposed here  is inspired on  the procedure used in the computation of the confidence level sets of the SR. 

Reformulation of classifiers to make them more robust or reliable is well established and defined. For example, the support vector machine (SVM) model (which we will show to be a scalable classifier) has been extensively studied (see \cite{Panchenko,Evgeniou,price,BENTAL19991,9408661}).

Recent approaches deal with finding regions with fixed false negative or positive rates, by means of sensitivity analysis of SVDD in \cite{carlevaro}, and of Boolean rules in \cite{SaraIS}, but they disregard the confidence interval of the resulting regions.

Understanding the concept of confidence more broadly, the following approaches can also be considered.
\textit{Out of distribution detection} \cite{NEURIPS2020_b90c4696}, where statistical tests are performed to assess if predictions are performed in conditions different from what was learned from the training stage.
\textit{Counterfactual eXplanation} \cite{9321372, 9787552}, which finds the minimum feature variation in order to change the predicted class.

In summary, the control of the confidence of the model~\cite{H_llermeier_2021} is the goal we pursue in this paper. The proposed methodology is suitable to address the potential variations in the underlying probability distribution of the data.

 The introduction of the scalable family of classifiers makes it possible to approximate the PSR while maintaining the same level of confidence. Together with probabilistic scaling, scalable classifiers provide a new framework for compliant machine learning.
What is new about this approach is that probabilistic assurance is provided in the design phase of the classifier, along with the calibration of the model. This is a key point because in the state of the art all methods that address the problem of providing probabilistic guarantees in prediction are based on a posteriori approaches. 

\subsection{Notation and order statistics concepts}

Given an integer $n$, $[n]$ denotes the integers from 1 to~$n$. 
Given $x\in\R$, $\lfloor x\rfloor$ denotes the greatest integer no larger than $x$ and $\lceil x\rceil$ the smallest integer no smaller than $x$. The set of non-negative reals is denoted $\R_+$.
%
%
Given integers $k,n$, and parameter $\varepsilon\in(0,1)$, the Binomial cumulative distribution function is denoted as
$$\Bin(k;n,\varepsilon) \doteq \Sum{i=0}{k}\binom{n}{i}\varepsilon^i(1-\varepsilon)^{n-i}.
$$
$\Pr\{A\}$ denotes the probability of the event $A$.
\\

The following definition is borrowed from the field of order statistics \cite{Alamo2018,MirasierraUQ}.

\begin{definition*}[Generalized Max]
Given a collection  of $n$ scalars $\Gamma = \{\gamma_i\}_{i=1}^n\in\R^n$,  and an integer $r\in [n]$,  we denote by
\[
\rmax{r}(\Gamma) 
\]
 the $r$-smallest value of $\Gamma$, so that there are no more than $r-1$ elements of $\Gamma$ strictly larger than
 $\rmax{r}(\Gamma) $. \end{definition*}

We will often refer to $r$ as a \textit{discarding parameter}, since the generalized max can be interpreted as a classical maximum after the largest $r-1$ points are discarded. Indeed, to construct $\rmax{r}(\Gamma) $ it is sufficient to order the elements of $\Gamma$ as
$\{\gamma_{(1)},
\gamma_{(2)},\ldots,\gamma_{(n)} \}$ so that 
$$
\gamma_{(1)}\ge\gamma_{(2)}\ge\cdots\ge\gamma_{(n)}.
$$ 
Then, we  let $\rmax{r}(\Gamma)  \doteq \gamma_{(r)}$.\\

The following result, see Property 3 in \cite{Alamo2018}, states how to obtain a probabilistic upper bound of a random scalar variable by means of the notion of generalized max.  This result has been used in the context of uncertainty quantification \cite{MirasierraUQ} and chance-constrained optimization  
\cite{Alamo:19:SafeApproximations,MAMMARELLA2022110108}, and plays a key role in our successive developments.

\begin{property}[Scaling factor \cite{Alamo2018}]\label{Property:Generalized:Max}
Given {\rm probabilistic parameters} $\varepsilon \in (0,1)$, $\delta\in (0,1)$ and a {\rm discarding parameter} $r\geq 1$, let $n\geq r$ be chosen such that 
\begin{equation}\label{ineq:Bin}\Bin(r-1;n,\varepsilon)
\leq \delta.
\end{equation}
Suppose that $\gamma\in\R$ is a random scalar variable with probability distribution $\mathcal{W}$. Draw $n$ i.i.d. samples $\{\gamma_i\}_{i=1}^{n}$ from distribution $\mathcal{W}$. Then, with a probability no smaller than $1-\delta$,
$$
\Pr_\mathcal{W}\left\{ \gamma > \rmax{r}(\{\gamma_i\}_{i=1}^n)\right\} \leq \varepsilon. $$\end{property}

In words, this results shows that, if the number of points is chosen large enough, the generalized max constitutes with very high probability a good approximation of the true maximum, in the sense that the probability of obtaining a larger value may be bounded a-priori.

The following corollary, proved in this form in \cite{MAMMARELLA2022110108}, provides a way to explicitly bound the number of samples $n$ (i.e.\ the so-called \textit{sample complexity}), by ``fixing" the discarding parameter
$r$ to be a percentage of the total number of samples, i.e.\ letting  $r=\beta\varepsilon n$.

\begin{corollary}[Explicit bound \cite{MAMMARELLA2022110108}]
\label{corollary:nice_ineq}
    Let $r = \left\lceil\beta\varepsilon n\right\rceil$, where $\beta \in (0,1)$, and define the quantity
    $$ \kappa \doteq \left(\frac{\sqrt{\beta} + \sqrt{2-\beta}}{\sqrt{2}(1-\beta)}\right)^2. $$
    Then, inequality \eqref{ineq:Bin} is satisfied for 
\begin{equation}
    \label{eq:utileinequality}
    n \ge \frac{\kappa}{\varepsilon}\ln\frac{1}{\delta}.
\end{equation}
Specifically, the choice $\beta = 0.5$ leads to $r = \displaystyle \left\lceil\frac{\varepsilon n}{2}\right\rceil$ and $\displaystyle n \ge \frac{7.47}{\varepsilon}\ln\frac{1}{\delta}$.
\end{corollary}

\section{Scalable Classifiers and Safe Sets}\label{section:ScalableClassifiers}
Classification algorithms assign to a given input vector $\x\in\mathcal{X}$ a single label $y\in\mathcal{Y}$ based on certain patterns in the data. We here consider the case of \textit{binary} classification, with classes  $y\in\{+1,-1\}$. Note that this choice of labels is without loss of generality, since any binary classifier can be converted to these labels. In our context, we assume that the label $+1$ denotes the target class $\mathrm{S}$, which is to be interpreted as ``safe" configurations.  The label $-1$ refers instead to the non-target class $\mathrm{U}$ (i.e. unsafe configuration). We write $\x\lbl \mathrm{S}$ to denote the fact that the vector $\x$ is safe, i.e.\ it has ``true" label $+1$. Similarly, $\x\lbl \mathrm{U}$ denotes that $\x$ is unsafe.

The main goal of our approach is to design, based on observations, a \textit{safety region}, i.e.\ a region $\mathcal{S}$ of the feature space $\mathcal{X}$ for which we have a guarantee that the probability of being unsafe 
is not larger than a given \textit{risk} level $\varepsilon\in(0,1)$.
More formally, we consider a probabilistic framework, and assume that the observations come from a fixed probability 
distribution. Then, for a given \textit{risk} level $\varepsilon\in(0,1)$, we are interested in constructing a  \textit{Probabilistic Safety Region} (PSR), denoted by $\mathcal{S}_\varepsilon$, satisfying 
\begin{equation}
\Pr\Bigl\{\x\lbl \mathrm{U}\text{ and } \x\in \mathcal{S}_\varepsilon\Bigr\} \le \varepsilon.
\end{equation}
In words, a PSR region $\mathcal{S}_\varepsilon\subseteq\mathcal{X}$ represents a set such that the probability of observing the event $\x\lbl \mathrm{U}$ conditioned to the event $\x\in  \mathcal{S}_\varepsilon$ is lower or equal than $\varepsilon$.

This paper provides a general approach to constructing such sets while maximizing a given performance index (e.g., size). We rely on a two-level of probability framework in which the final provided set is a probabilistic safety region with a (prescribed) level of probability.
To generate such probabilistic safety regions, we introduce a special (but rather general) class of classifiers, that we refer to as \textit{scalable classifiers} (SCs). Namely, SCs are classifiers whose formulation can be made to explicitly depend on a \textit{scaling parameter} $\rho\in\R$. 
The parameter $\rho$ allows to dynamically adjust the boundary of the classification: changing~$\rho$ causes the classifier to widen, shrink, or change shape completely. With some abuse of notation, we can think of the different values of $\rho$ as different ``level sets" of the classification function.

\subsection{Scalable Classifiers}
\label{subseq:scalable_class}

Formally, we consider binary classifiers which can be formulated as follows
\begin{equation}
\label{eq:phi:minus}
     \phi_{\thB}(\x,\rho)  \doteq 
    \begin{cases}
        +1 \quad \quad \text{if } \, f_\thB(\x,\rho) < 0, \\
        -1 \quad \quad \text{otherwise.}
    \end{cases}
\end{equation}
where  the function $f_\thB: \mathcal{X}\times\R \longrightarrow \R$ is the so-called \textit{classifier predictor}.
Note that, in the above notation, we highlight the fact that $f_\thB$ may depend also on a second set of parameters 
$\thB\subset\R^{n_{\thB}}$. The vector $\thB=[\thB_1,\cdots,\thB_{n_{\thB}}]^\top$ collects the so-called classifier \textit{hyperparameters}, that is all those 
parameters to be adjusted in the model (e.g.\ different choices of kernel,  regularization parameters, etc.). Obviously, a different choice of  $\thB$ corresponds to a possibly very different classifier. 
The role  of different choices of $\thB$ in the construction of the classifier is extremely important, and will be discussed in Section \ref{sec:families}.\\

Note that $\phi_{\thB}(\x,\rho)$ may be interpreted as the ``tentative" (or predicted) label associated to the point $\x$ by the classifier.
In the sequel, with some slight abuse of notation, we will sometimes refer to $f_{\thB}$ as the classifier itself. 

As we will see in the following derivations, the scaling parameter $\rho$ plays a key role for this particular family of classifiers. Specifically, in order to define a SC, we require that the $\rho$-parametrized classifier $\phi_{\thB}(\x,\rho)$ satisfies some special ``ordering" condition, as defined next.

\begin{assumption}[Scalable Classifier]\label{assum:Conditions:on:f} We assume that for every $\x \in \mathcal{X}$, $f_{\thB}(\x,\rho)$ is a continuous and monotonically increasing function on $\rho$, i.e. 
\begin{equation}\label{eq:increasing}
\rho_1 >\rho_2\; \Rightarrow  f_{\thB}(\x,\rho_1) > f_{\thB}(\x,\rho_2), \quad\forall \x\in\mathcal{X}.
\end{equation}
We assume also that
\begin{equation} \label{eq:limits:f:prop}
\lim\limits_{\rho\to -\infty} f_{\thB}(\x,\rho) < 0 < \lim\limits_{\rho\to \infty} f_{\thB}(\x,\rho), \quad \forall \x \in \mathcal{X}. \end{equation}
\end{assumption}

\begin{property}[Boundary radius]\label{prop:existence:uniqueness}
Suppose that Assumption \ref{assum:Conditions:on:f} holds. Then, for each $\x\in \mathcal{X}$, there exists a unique $\bar{\rho}(\x)$ satisfying $f_\thB(\x,\bar{\rho}(\x))=0$. Moreover, the classifier $\phi_{\thB}(\x,\rho)$ given by \eqref{eq:phi:minus} satisfies 
$$ \phi_\theta(\x,\rho) = -1  \Leftrightarrow \rho \geq \bar{\rho}(\x).$$
\end{property}
The proof of Property \ref{prop:existence:uniqueness} is available in Appendix \ref{proof: property1}. In this paper, and under Assumption \ref{assum:Conditions:on:f}, we denote $\bar{\rho}(\x)$ the unique solution (see Property \ref{prop:existence:uniqueness}) to the equation 
$$ f_\thB(\x,\rho) = 0.$$
In words, a scalable classifier is a classifier for which, given $\x$, there is always a value of $\rho$, denoted $\bar{\rho}(\x)$, that establishes the border between the two classes. 
Therefore, a SC is a classifier that maintains the target class of a given feature vector $\x$ under a decrease of $\rho$ 
We also remark that this definition is implied by condition \eqref{eq:increasing}.  Indeed, for a given $\tilde\x \in \mathcal{X}$ and $\rho_1>\rho_2$, if $f_{\thB}(\tilde\x,\rho_1) < 0$ (i.e. $\phi_{\thB}(\tilde\x,\rho_1) = +1 $) then $f_{\thB}(\tilde\x,\rho_2) < f_{\thB}(\tilde\x,\rho_1) < 0$ (i.e. $\phi_{\thB}(\tilde\x,\rho_2) = +1$). 

Next property shows that any standard binary classifier can be rendered scalable by simply including the scaling parameter $\rho$ in an additive way.
\begin{property}\label{prop:SC:through:adding:rho}
Consider the function $\hat{f}_\thB:\mathcal{X}\to \R$ and its corresponding classifier 
$$      \hat{\phi}_{\thB}(\x)  \doteq 
    \begin{cases}
        +1 \quad \quad \text{if } \, \hat{f}_\thB(\x) < 0, \\
        -1 \quad \quad \text{otherwise.}
    \end{cases}
    $$
Then, the function $f_\thB (\x,\rho) = \hat{f}_\thB(\x)+\rho$ satisfies Assumption \ref{assum:Conditions:on:f} and thus provides the scalable classifier 
$$ \phi_{\thB}(\x,\rho)  \doteq 
    \begin{cases}
        +1 \quad \quad \text{if } \, f_\thB(\x,\rho) < 0, \\
        -1 \quad \quad \text{otherwise.}
    \end{cases}
    $$
\end{property}
\begin{proof} The result is trivial because $f_\thB (\x,\rho) = \hat{f}_\thB(\x)+\rho$ is clearly a continuous and monotonically increasing function on $\rho$.
It is also straightforward to check that \eqref{eq:limits:f:prop} is satisfied.\end{proof}
The next example illustrates the use of the previous property to obtain a scalable classifier from a standard linear classifier. 
In subsection \ref{Sub:sec:Examples}  we present other examples of scalable classifiers and we show that the scaling parameter $\rho$ does not need necessarily to appear in an additive way into $f_\thB(\x,\rho)$ to obtain a SC.

\begin{example}[Linear classifier as scalable classifier] 
\label{ex:linear_classifier}
Consider the standard linear classifier defined by means of the function 
\begin{equation*}
\label{eq:lin}
    \hat{f}_{\thetaB}(\x) = \w^\top\x - b.
\end{equation*}
The classifier elements $\w,b$ may be obtained, for instance, as the solution of a SVM problem of the form
\[
\min\limits_{\w,b}  \frac{1}{2\eta}\w^\top \w + \frac{1}{2}\Sum{i=1 }{n} 
\max\left\{
0,1+y_i(\w^\top \varphi(\x_i) - b)
\right\},
\]
and notice that we are not using the usual form of the hinge loss with a minus in front of $y_i$ ($\max\left\{
0,1-y_i(\w^\top \varphi(\x_i) - b)
\right\}$) since we would like $\hat{f}_\thB(\x)$ to be negative for $y=+1$ and positive otherwise.\\
In this case, the classifier depends on the choice of the regularization term $\eta$, and of the specific regressor functional $\varphi(\cdot)$. That is, for a fixed choice of regressor, the hyperparameter vector is just the scalar $\thetaB=\eta$. In this sense, we remark that a more rigorous notation would be
 $\w = \w(\thetaB)$ and $b = b(\thetaB)$, but we omit this dependence for the sake of readability.
 
As stated in the proof of Property \ref{prop:SC:through:adding:rho}, it is immediate to observe that  linear classifiers belong indeed to the class of scalable classifiers if we introduce a scaling parameter $\rho$ in an additive way, that is
\begin{equation}
        f_{\thetaB}(\x,\rho) = \w^\top\x - b + \rho.
\end{equation}
Indeed, given $\rho_1 > \rho_2$ we immediately have that 
\begin{equation*}
     \w^\top\x - b+\rho_1 > \w^\top\x - b+\rho_2, \quad \forall \x\in \mathcal{X},
\end{equation*}
and it is straightforward to see that also \eqref{eq:limits:f:prop} holds.
\end{example}

\subsection{Main Result: Probabilistic Safety Regions}
\label{subseq:PSR}

Consider a given SC classifier $f_\thB(\x,\rho)$ (i.e.\ a classifier designed considering a specific choice\footnote{Note that in this section, we assume $\thB$ to be fixed and given. Section~\ref{sec:families} discusses in detail how the possibility of choosing $\thB$ may be exploited to improve the SC.} of hyperparameter~$\thB$). Then, for a given value of the scaling parameter $\rho\in\R$, we define the \textit{$\rho$-safe set}
$$ \mathcal{S}(\rho) = \set{\x\in\mathcal{X} }{f_\thB(\x,\rho)<0},$$
which represents the set of points $\x\in\mathcal{X}$ predicted as safe by the classifier with the specific choice $\rho$ 
 i.e. the safety region of the classifier $f_\thB$ for given $\rho$.

{Note that, contrary to the similar concept introduced in \cite{MAMMARELLA2022110108}, the interpretation of $\rho$ as a ``radius" is not valid anymore. That is,  $\rho$ shall be viewed as a \textit{safety inducing parameter}: the larger $\rho$, the more stringent the requirements will be and thus, the smaller will be the corresponding $\rho$-safe set. Also, note that $\rho\in\R$, hence it can take negative values.  \footnote{One may salvage the ``radius" interpretation by introducing a new parameter $\tilde\rho\doteq - \rho$, but we prefer not to do this because this choice would complicate the ensuing derivations.}.
The reason why we prefer the current formulation is that it allows us to draw a clear parallel with the results in conformal prediction, see Remark \ref{remark:CP}.
Indeed, following the reasoning above, it is easy to see that the larger $\rho$, the \textit{smaller} is the region. In other words, it is easy to see that 
\begin{equation*}
\rho_1 > \rho_2 
\Longrightarrow \mathcal{S}(\rho_1) \subset \mathcal{S}(\rho_2). 
\end{equation*}
This behavior is depicted in Fig.\ \ref{Fig:SC}. 
\begin{figure}[!h]
\centering
\includegraphics[width=6cm]{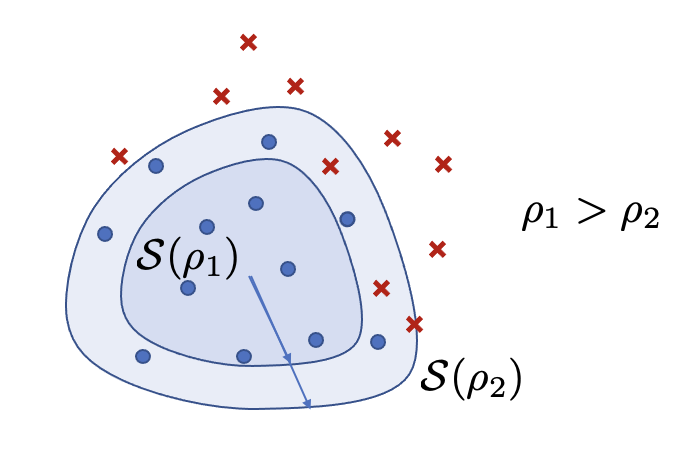}
\caption{Graphical depiction of the role of the scaling parameter. The blue circles represent safe points $\x\lbl \mathrm{S}$, while the red crosses represent unsafe ones, $\x\lbl \mathrm{U}$.}
\label{Fig:SC}
\end{figure}

In the next section, for completeness, we  present some notable examples of well-assessed classifiers which can be reformulated in a way so that  they belong to the SC family. 
We now introduce the main result of the paper, which is a simple procedure to obtain, from a calibration set $\mathcal{Z}_c \doteq \left\{(\x_i,y_i)\right\}_{i=1}^{n_c}$, a probabilistic safety region $\mathcal{S}_\varepsilon$ that with a probability no smaller than $1-\delta$ satisfies the probability constraint
$$ \Pr\Bigl\{y=-1 \text{ and } \x \in \mathcal{S}_\varepsilon\Bigr\}  \leq  \varepsilon.$$
We will assume that the pair $(\x,y)$ is a random variable and that $\Pr\{\x\in \mathcal{X}\}=1$. Moreover, the $n_c$ samples of $\mathcal{Z}_c$ are assumed i.i.d.. 
\begin{theorem}[Probabilistic Safety Region]
\label{theorem:main}
Consider the classifier \eqref{eq:phi:minus}, and suppose that Assumption \ref{assum:Conditions:on:f} holds and that  $\Pr\{\x\in \mathcal{X}\}=1$.  
Given a calibration set $\mathcal{Z}_c \doteq \left\{(\x_i,y_i)\right\}_{i=1}^{n_c}$ ($n_c$ i.i.d. samples), suppose that $\delta\in (0,1)$, $\varepsilon\in (0,1)$, and the integer \textit{discarding parameter} $r$ satisfies $n_c\ge r \geq 1$, and
$$\Bin(r-1;n_c,\varepsilon)
\leq \delta.$$ 
Consider the subset $\mathcal{Z}_c^U= \left\{(\tilde{\x}^U_j,-1)\right\}_{j=1}^{n_U} $ corresponding to all the unsafe samples in $\mathcal{Z}
_c$ and define the {\rm probabilistic scaling of level} $\varepsilon$ as follows
\begin{equation}
    \label{eq:rhoeps}
\rho_\varepsilon \doteq
\mathrm{max}^{(r)}\left(\{\bar{\rho}(\tilde{\x}^U_{j})\}_{j=1}^{n_U}\right),
\end{equation}
and define the corresponding $\rho_\varepsilon$-safe set
\begin{eqnarray*} 
\mathcal{S}_\varepsilon & \doteq & \bsis{cc} \mathcal{S}\left(\rho_\varepsilon\right)  & \text{ if }  n_U\geq r \\ \mathcal{X} & \text{otherwise.} \esis
\end{eqnarray*} 
Then, with probability no smaller than $1-\delta$,  
\begin{equation}
\Pr\Bigl\{y=-1 \text{ and } \x \in \mathcal{S}_\varepsilon\Bigr\}  \leq \varepsilon.
\label{eq:theorem}
\end{equation}
\end{theorem}
The proof of Theorem \ref{theorem:main} is available in Appendix \ref{proof:teo-main}.\\
From  Corollary \ref{corollary:nice_ineq}, it is easy to see that the smaller $\varepsilon$ is, the larger $\mathrm{max}^{(r)}$ is (i.e., $\rho_\varepsilon$).

In this view, $\rho_\varepsilon$ is in line with the idea of a safety-inducing parameter discussed previously, since it operationally encodes the probabilistic guarantee required by $\varepsilon$. 

\subsection{Examples of Scalable Classifiers}\label{Sub:sec:Examples}

As stated by Property \ref{prop:SC:through:adding:rho}, any standard classifier can be converted into a scalable one by means of the inclusion of the scaling parameter $\rho$ in an additive way. Using this scheme, we present in this section families of scalable classifiers obtained from Support Vector Machines and Support Vector Data Description classifiers. Additionaly, we also show how to obtain a scalable classifier from the Logistic Regression classifier by including the parameter $\rho$ in a non-additive manner. 
In the following examples, we assume we are given a learning set 
\[
\mathcal{Z}_L\doteq\left\{\left(\x_i,y_i\right)\right\}_{i=1}^n \subseteq \mathcal{X}\times\left\{-1,+1\right\}
\]
containing observed feature points and corresponding labels $\z_i=\left(\x_i,y_i\right)$.
Then, we introduce the kernels (see e.g.~\cite{Hofmann_2008}).
In particular, letting 
$$\varphi : \mathcal{X} \longrightarrow \mathcal{V}$$ 
be a \textit{feature map} (where $\mathcal{V}$ is an inner product space) we define
\begin{eqnarray} 
\Phi &=& \bmat{cccc} \varphi(\x_1) & \varphi(\x_2) & \ldots & \varphi(\x_n)\emat, \label{not:R} \\ 
D&=&\text{diag}\{ y_1, y_2,\ldots, y_n\} \label{not:D}, \\
K&=&\Phi^\top\Phi \label{not:K},
\end{eqnarray}
 with $ K_{i,j} = K(\x_i,\x_j) = \varphi(\x_i)\T \varphi(\x_j), \; i\in[n], j\in[n]$ the kernel matrix.\\
The models considered and their derivation is absolutely classical. However, since we are interested in scalable classifiers with guaranteed safety, for each model we will consider two hyperparameters, i.e.\ we will set 
$\thetaB = [ \eta ,\, \tau ]^\top$, where besides the classical regularization parameter $\eta\in\R$  we introduce a weighting term  $\tau\in(0,1)$ that penalizes missclassification errors (the role of $\tau$ is much in the spirit of quantile regression formulation \cite{koenker_2005}).

\subsubsection{Scalable SVM}

SVM is the simplest extension of a linear model and indeed we define its classifier predictor as
\begin{equation*}
    \hat{f}_\thB(\x) = \w^\top\varphi(\x) - b.
\end{equation*}

The SVM formulation we adopt is the classical one proposed by Vapnik in \cite{10.1023/A:1022627411411}, with the addition of the weighting  parameter $\tau$:
%
\begin{eqnarray}
 &\min\limits_{\w,b , \xi_1,\ldots, \xi_n } & \frac{1}{2\eta}\w^\top \w + \frac{1}{2}\Sum{i=1 }{n} \left((1-2\tau)y_i +1\right)\xi_i\nonumber  \\
& \text{s.t.} & y_i(\w^\top \varphi(\x_i) - b) \leq \xi_i-1, \; i\in[n],\label{eq:SVM}\\ 
& &\xi_i \geq 0, \; i\in[n].\nonumber
\end{eqnarray}
We explicitly report the formulation in \eqref{eq:SVM} since our specific definition of the classifier, which requires $f_\thB(\x_i) $ to be negative when $y_i$ equals +1, leads to a slightly different formulation than the classical one.

The offset $b$ can be found exploiting special feature points $\x_s$ called \textit{support vectors} that are such that $\varphi(\x_s)$ lies on the boundary of the transformed space.
The addition of the scaling parameter $\rho$ changes the model in
\begin{eqnarray}
\label{eq:classpredSVM}
f_\thB(\x,\rho) &=& \w^\top\varphi(\x) - b + \rho.
\end{eqnarray}

We observe that, for the linear kernel, the variation of $\rho$ is simply a rigid translation of the classification hyperplane; for other kernels, for example, the Gaussian kernel or the polynomial kernel, the effect is the ``deflation'' or the ``inflation'' of the classification boundary. The composition with the feature map does not affect the scalability property of the linear classifier, so it is easy to verify from the considerations made in \ref{ex:linear_classifier} that indeed scalable SVM satisfies Assumption \ref{assum:Conditions:on:f} (see also Property \ref{prop:SC:through:adding:rho}).

\begin{remark}[On the role of $\tau$ parameter.]
\label{rem:tau}
%
%
%
%
%
%
Indeed, it is easy to see that small values of $\tau$ add more weight to the class $+1$, which is the class we are interested in. So, the choice of a ``good" value of $\tau$ is particularly important. This will be discussed in Section \ref{sec:families}, where the possibility of considering several values for this parameter in the context of our approach is discussed in detail.
\end{remark}
\subsubsection{Scalable SVDD}

\begin{figure*}[!t]
\centering
\subfloat[Scalable SVM]{
\includegraphics[width=2.5in]{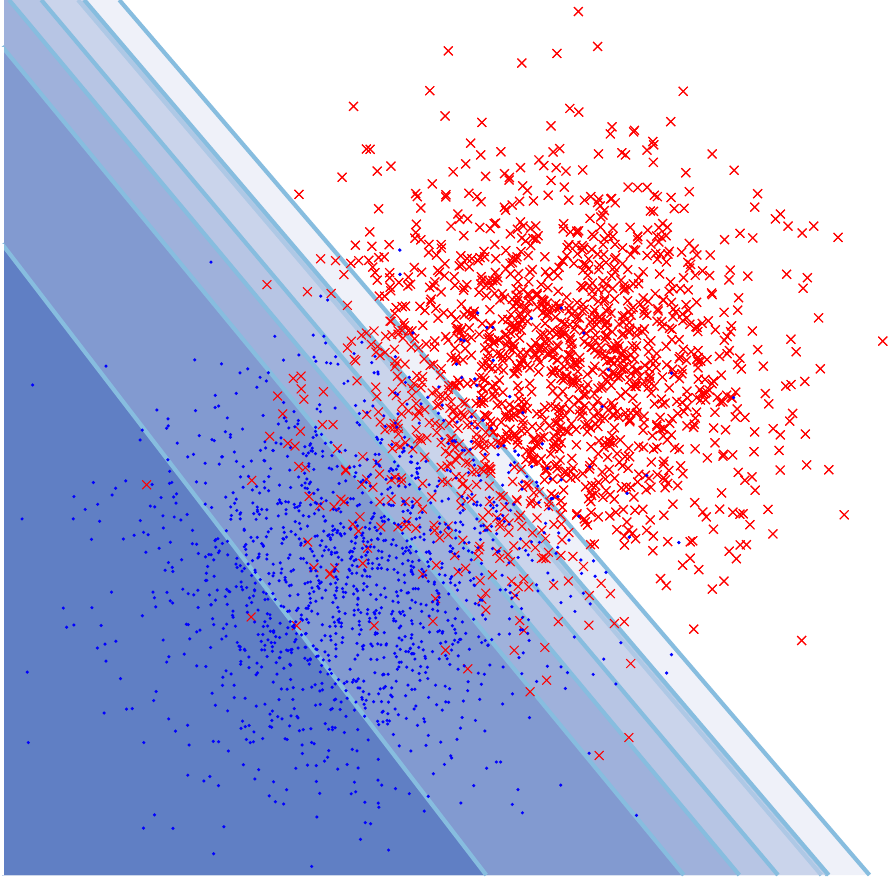}
\label{fig:SCexampleA}
}
\subfloat[Scalable SVDD]{\includegraphics[width=2.4in]{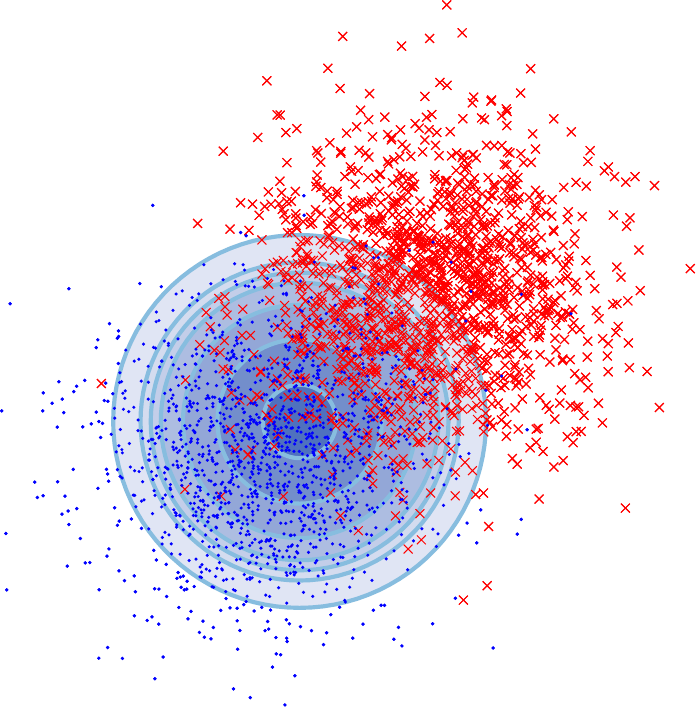}
\label{fig:SCexampleB}
}
\subfloat[Scalable LR]{\includegraphics[width=2.4in]{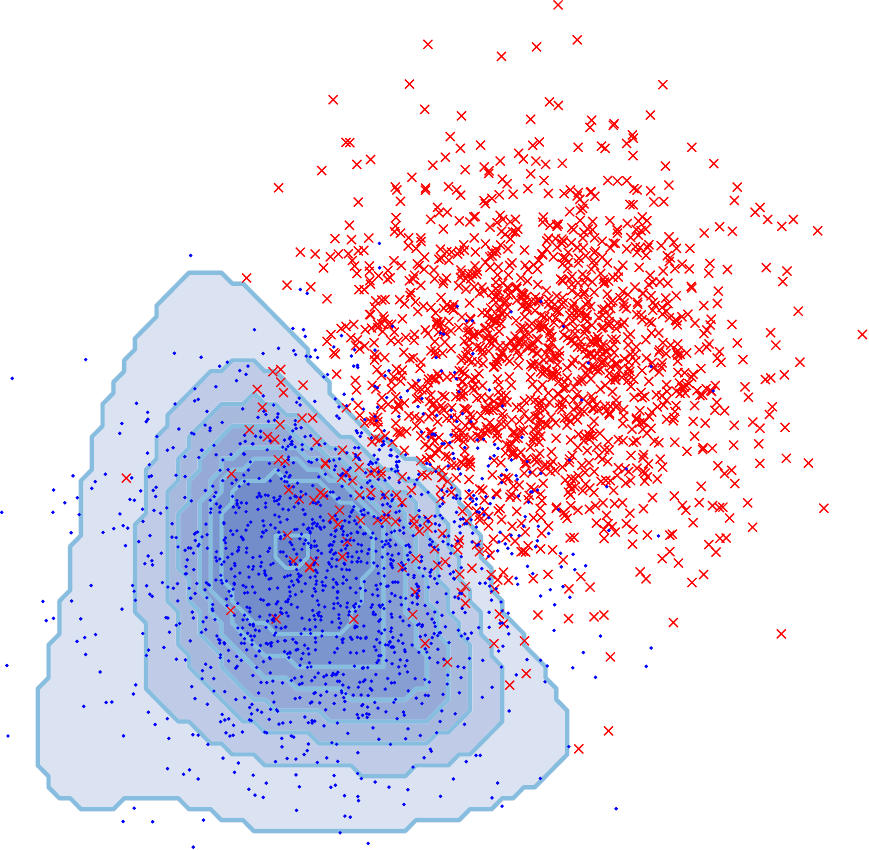}
\label{fig:SCexampleC}}
\vfill
\subfloat{\includegraphics[width=2.4in]{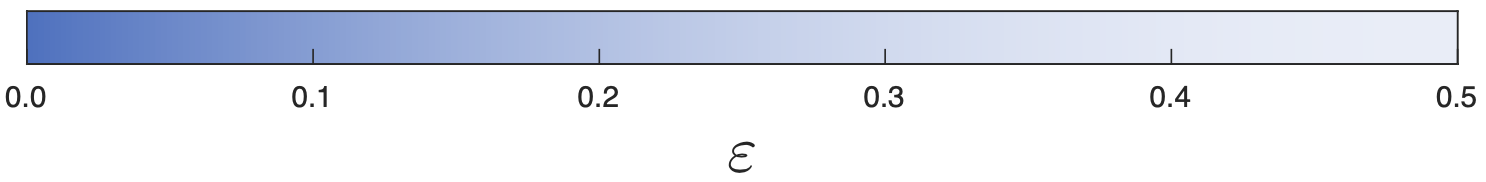}}
\caption{2D examples of PSRs via, respectively from left to right, scalable SVM (linear kernel), scalable SVDD (linear kernel) and scalable LR (Gaussian kernel). Synthetic test data were sampled from Gaussian distributions and classified for varying values of $\varepsilon$ (from $0.01 \ \text{to} \ 0.5$, from lighter to darker colors), after calibrating the scalable parameters with a calibration set of size $n_c$ according with bound \eqref{eq:utileinequality} and $\delta = 10^{-6}$.  Blue points refer to the safe class ($\x\lbl \mathrm{S}$) and reds to the unsafe one ($\x\lbl \mathrm{U}$).}
\label{fig:SCexample}
\end{figure*}

SVDD was introduced in \cite{SVDD} based on the idea of classifying the feature vectors by enclosing the target points (in the kernel space) in the smallest hypersphere of radius $R$ and center  $\w$. With this idea, we define the scalable classifier predictor for SVDD as
\begin{equation}
\label{eq:SVDDclasspred}
    f_{\thetaB}(\x,\rho) = \norm{\varphi(\x)-\w}^2 - (R^2 - \rho),
\end{equation}
where $\w,R$ are obtained as the solution of the following weighted optimization problem
\begin{eqnarray}  &\min\limits_{\w, R, \xi_1,\dots,\xi_n} & \frac{1}{2\eta}R^2 + \frac{1}{2}\Sum{i=1}{n} \left((1-2\tau)y_i+1\right)\xi_i  \label{eq:SVDD}\\
& \text{s.t.}& y_i\left(\norm{\varphi(\x_i)-\w}^2-R^2\right) \leq \xi_i, \; i\in[n],\nonumber\\ 
& &\xi_i \geq 0, \; i\in[n]\nonumber
\end{eqnarray}
that, again, depends on the hyperparameters $\thetaB = [ \eta ,\, \tau ]^\top$, playing the role of regularization and missclassification parameters. As for the SVM model, the radius $R$ is retrieved by support vectors, that are feature points lying on the hypersphere boundary of the classification in the kernel space. 
 It is immediate to observe that the introduction of the scaling parameter $\rho$ maintains the idea that an increase in $\rho$ will result in a smaller radius $\tilde{R}^2 = {R^2-\rho}$, thus implying the meaning of safety induction}. 
Indeed, the scalable SVDD-classifier predictor \eqref{eq:SVDDclasspred} clearly satisfies equations \eqref{eq:increasing} and~\eqref{eq:limits:f:prop}.

\subsubsection{Scalable Logistic Regression}

Logistic Regression (LR) classifies points $\x\in\mathcal{X}$ on the basis of the probability expressed by the logistic function 
%
\begin{eqnarray*}
    \frac{1}{1+e^{\w^\top\varphi(\x)-b}} &=&  \Pr_\thB\left\{y=+1\mid\x\right\} \\
     &=& 1-\Pr_\thB\left\{y=-1\mid\x\right\},
\end{eqnarray*}
where $\w$ and $b$ minimize the regularized negative log-likelihood
\begin{eqnarray*}
   && \mathrm{L}\left(\w,b \mid \x,y\right)=\dfrac{1}{2\eta}\w^\top\w \\
    &&+\frac{1}{2}\Sum{i=1}{n}\left((1-2\tau)y_i+1\right)\text{log}\left(1+e^{y_i\left(\w^\top\varphi(\x_i)-b\right)}\right),
\end{eqnarray*}
with $b$ explicitly computed with the support vectors of the model. Note that, differently from classical LR and in the spirit of previously described approaches, we introduce into the cost function the weight parameter $\tau\in(0,1)$ to penalize misclassification, we also consider the regularization parameter $\eta>0$. In this case, defining 
$$ \hat{f}_{\thetaB}(\x) = \frac{1}{2}-\frac{1}{1+e^{\w^\top\varphi(\x)-b}}, $$
the standard LR classifier is given by
$$  \hat{\phi}_{\thB}(\x)  \doteq 
    \begin{cases}
        +1 \quad \quad \text{if } \, \hat{f}_\thB(\x) < 0, \\
        -1 \quad \quad \text{otherwise.}
    \end{cases}
    $$ 
We now show that the following function
$$  f_{\thetaB}(\x,\rho) = \frac{1}{2}-\frac{1}{1+e^{\w^\top\varphi(\x)-b + \rho, }} $$
satisfies Assumption \ref{assum:Conditions:on:f}, and thus provides a scalable classifier. 
Clearly,  $f_{\thetaB}(\x,\rho)$ is a continuous and monotonically increasing function on $\rho$. Moreover, 
$$\lim_{\rho \to -\infty}f_\theta(\x,\rho) = -1/2 < 0 < 1/2 = \lim_{\rho \to +\infty}f_\theta(\x,\rho). $$
Thus, we conclude that $$ \phi_{\thB}(\x,\rho)  \doteq 
    \begin{cases}
        +1 \quad \quad \text{if } \, f_\thB(\x,\rho) < 0, \\
        -1 \quad \quad \text{otherwise.}
    \end{cases}
    $$
is a scalable classifier.



\begin{remark}[Generality of SC]
    We remark that the three examples above, although already significant in themselves, represent only a small subset of possible scalable classifiers.
    Indeed, as stated in Property \ref{prop:SC:through:adding:rho}, any standard classifier can be easily converted into an scalable one. Thus, the results presented in this paper can be directly applied, for example, to any deep neural network classifier. 
\end{remark}

\begin{remark}
We emphasize that one of the main advantages of our approach is that the distribution of the calibration set need not be equal to that of the learning set. It should be equal to the one for which we want to impose probabilistic guarantees. This is a crucial observation, since probabilistic guarantees apply only to the distribution from which the calibration set was drawn, which must therefore be chosen carefully. Note also that as the desired degree of guarantee changes, the cardinality required for the the calibration set changes. 
\end{remark}
\begin{example}
\label{ex:GaussianPSR}
To give the reader a simple but meaningful idea of the method, Figure \ref{fig:SCexample} shows the behavior of the PSR as $\varepsilon$ varies while $\delta$ is fixed to $10^{-6}$. For this example, we sampled with equal probability two classes, ``safe'' $\mathrm{S}$ and ``unsafe'' $\mathrm{U}$, from two Gaussian distributions with respectively means and covariance matrices
$$\mu_\mathrm{S} = \begin{bmatrix}
-1 \\
-1
\end{bmatrix}, \, \Sigma_\mathrm{S} = \mathrm{I} \, \, ; \, \,
\mu_\mathrm{U} = \begin{bmatrix}
+1 \\
+1
\end{bmatrix}, \, \Sigma_\mathrm{U} = \mathrm{I} $$
where $\mathrm{I}$ is the identity matrix. We sampled $3000$ points for the training set and $10000$ for the test set, and $n_c = n_c(\varepsilon)$ points for the calibration set according to Corollary \ref{corollary:nice_ineq} (from $146$ points for $\varepsilon = 0.5$ up to $7261$ for $\varepsilon = 0.01$).\\
The behaviour of the PSR constructed via the scalable classifiers is in agreement with the theory developed: the smaller the $\varepsilon$ (i.e. the smaller is the error required) the smaller is the PSR, to guarantee more probability of safety. For scalable SVM (left) and scalable SVDD (middle) we choose a linear kernel, while for scalable LR (right) a Gaussian kernel was used. The blue cirles represent safe points $\x\lbl \mathrm{S}$, while the red crosses represent unsafe ones, $\x\lbl \mathrm{U}$.
\end{example}
\section{Finite families of hyperparameters}
\label{sec:families}

Probabilistic scaling guarantees confidence in prediction for any given scalable classifier. In other words, for any \textit{fixed} value of hyperparameter $\boldsymbol{\theta}$, the  safety set obtained selecting the scaling parameter $\rho$ according to our procedure will fulfill the required probabilistic guarantees (Theorem \ref{theorem:main}). 
However, it should be remarked that different values of $\boldsymbol{\theta}$ will correspond to different models, and the resulting set will consequently be different, both in ``size" and in ``goodness". In particular, if the starting SC has been chosen badly, our procedure would lead to a very small PSR, that would be indeed guaranteed theoretically, but with no practical use.

Hence, the problem of selecting the best initial SC becomes of great importance. In our setup, this problem translates in choosing the best value for the hyperparameter. Also, we remark that, in general,
there may be other parameters that affect the performance of a classifier, such as the choice of different kernels or different weights or different regularizations and many others. Hence, in general, the hyperparameter $\thetaB$ may be of larger dimensions and consider several possible choices. 
\\
To formally state our problem, we assume to have a finite set of $m$ possible hyperparameters to choose from
\begin{equation}
\label{eq:Theta} \Theta = \left\{\thetaB^{(1)}, \thetaB^{(2)}, \dots, \thetaB^{(m)}\right\}, 
\end{equation}
and we consider the problem of selecting the ``best" one.

Hence, we assume we are given a performance function $J:\Theta\to \R$ which  measures the goodness of the model described by~$\boldsymbol{\theta}$. 
Then, we will choose 
\[
\thetaB^\star \doteq \arg\max_{\thetaB\in\Theta}J(\thetaB).
\]

Clearly, depending on the problem at end, different cost functions may be devised. We discuss a possible meaningful choice of performance function in Section \ref{sec:miss}. In the following section, we show how the scaling procedure can be easily modified to guarantee that the selected SC, and the ensuing estimate of the PSR, still enjoy the desired probabilistic guarantees.

\subsection{Probabilistic scaling for finite families of SC}

The following results, whose proof is a direct consequence of Bonferroni's inequality and is omitted for brevity, shows how the results in Theorem~\ref{theorem:main}
may be immediately extended to the case of a finite family of classifiers, i.e. a finite set of candidate SCs described by a finite set of possible values of hyperparameters. 
\begin{figure*}[!t]
\centering
\subfloat[$\eta = 10^{-2}, \tau = 0.1$ \\ $J(\eta,\tau) = 535 $ ]{\includegraphics[width=2in]{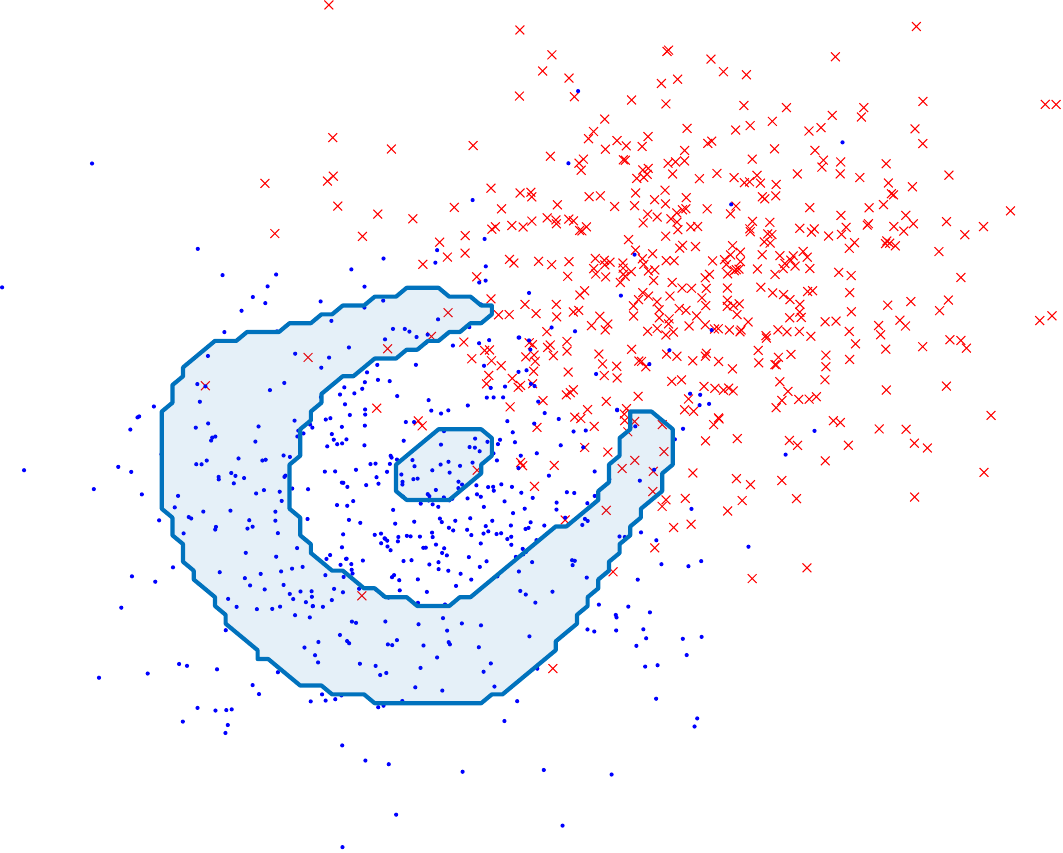}
\label{Fig:FF00101}}
\hfil
\subfloat[$\eta = 10^{-1}, \tau = 0.1$ \\ $J(\eta,\tau) = 563$]{\includegraphics[width=2in]{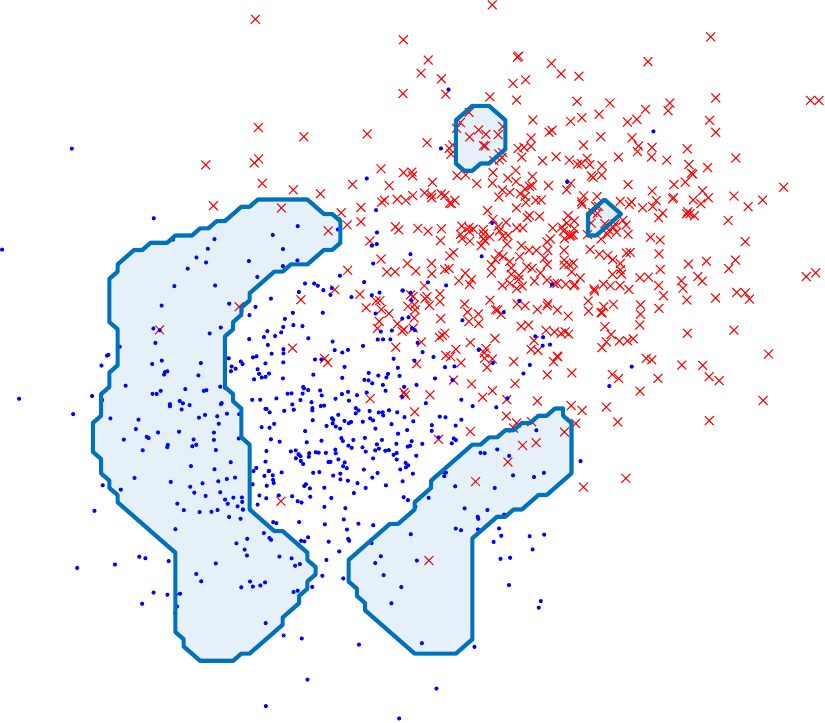}
\label{Fig:FF0101}}
\hfil
\subfloat[$\eta = 1, \tau = 0.1$ \\ $J(\eta,\tau) = 590$]{\includegraphics[width=2in]{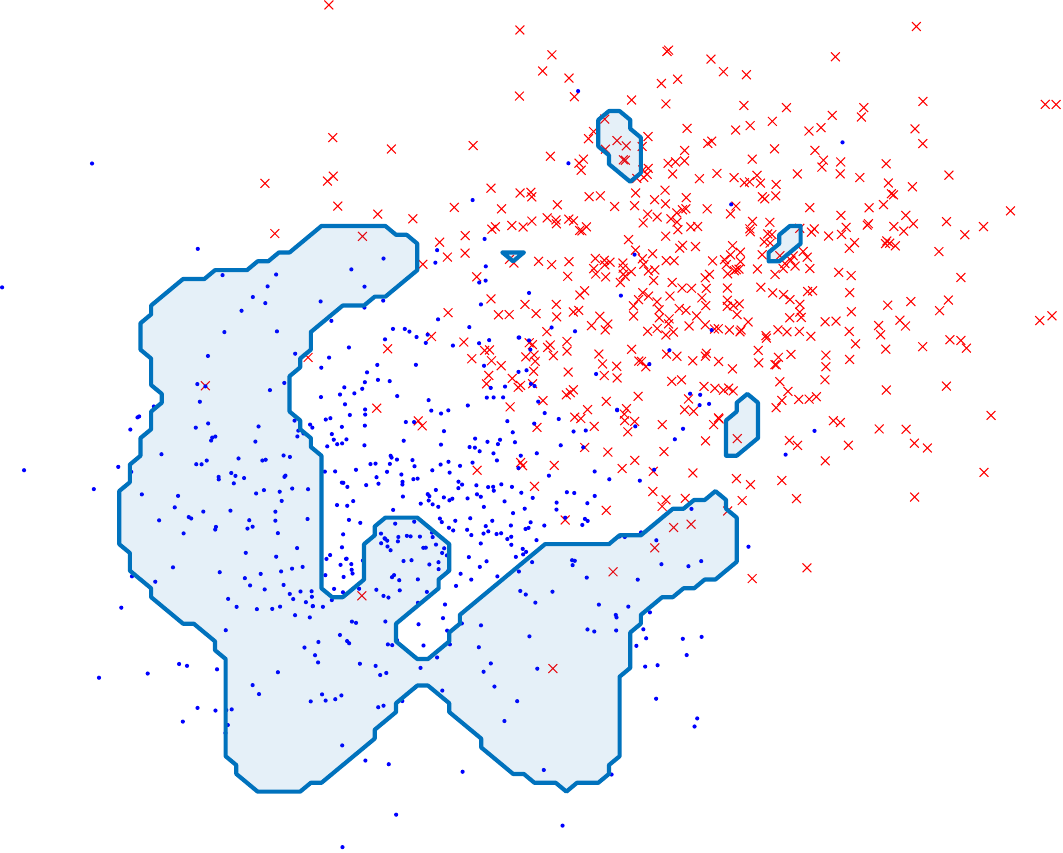}
\label{Fig:FF101}}
\vfil
\subfloat[$\eta = 10^{-2}, \tau = 0.5$ \\ $J(\eta,\tau) = 697$]{\includegraphics[width=1.9in]{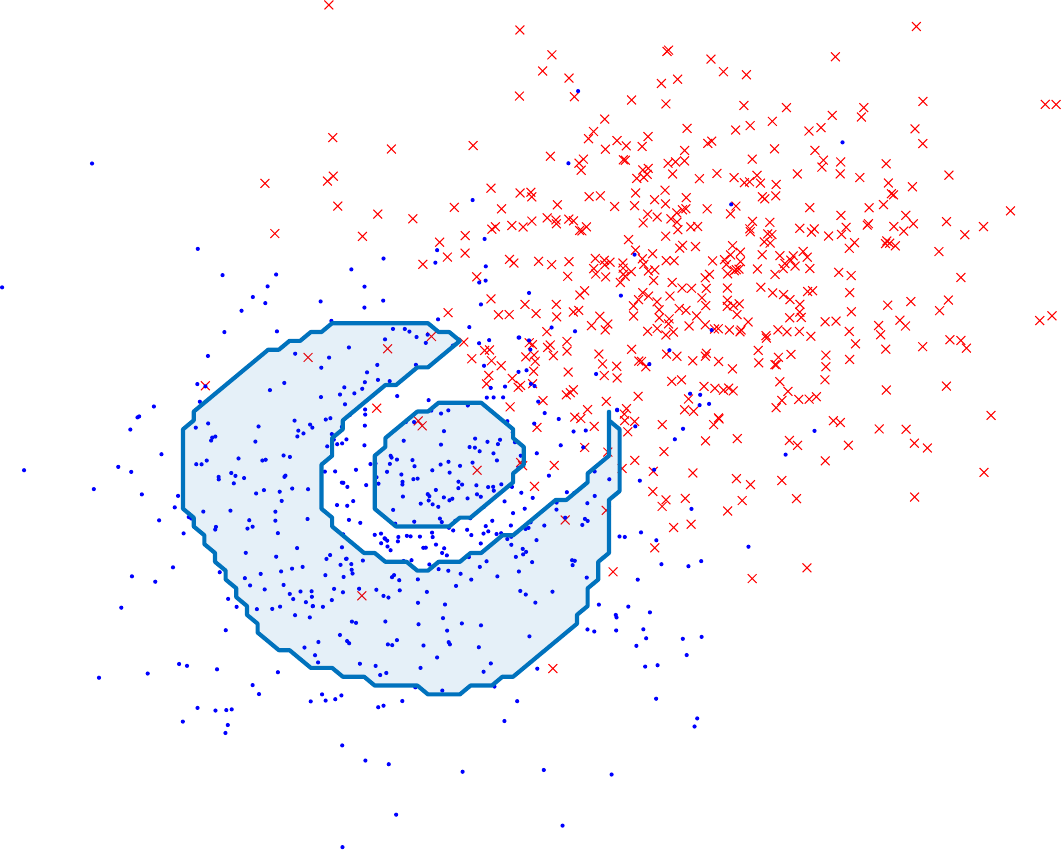}
\label{Fig:FF00105}}
\hfil
\subfloat[$\eta = 10^{-1}, \tau = 0.5$ \\ $J(\eta,\tau) = 776$]{\includegraphics[width=2in]{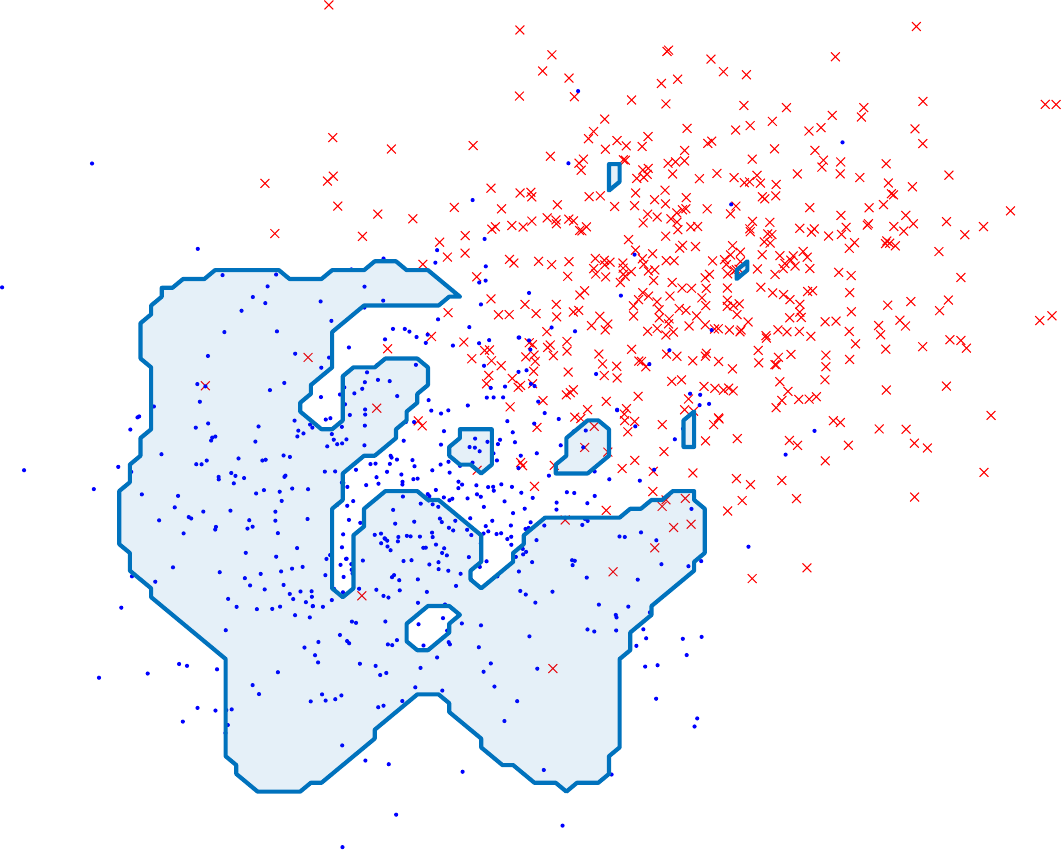}
\label{Fig:FF0105}}
\hfil
\subfloat[$\eta = 1, \tau = 0.5$ \\ $J(\eta,\tau) = 736$]{\includegraphics[width=2in]{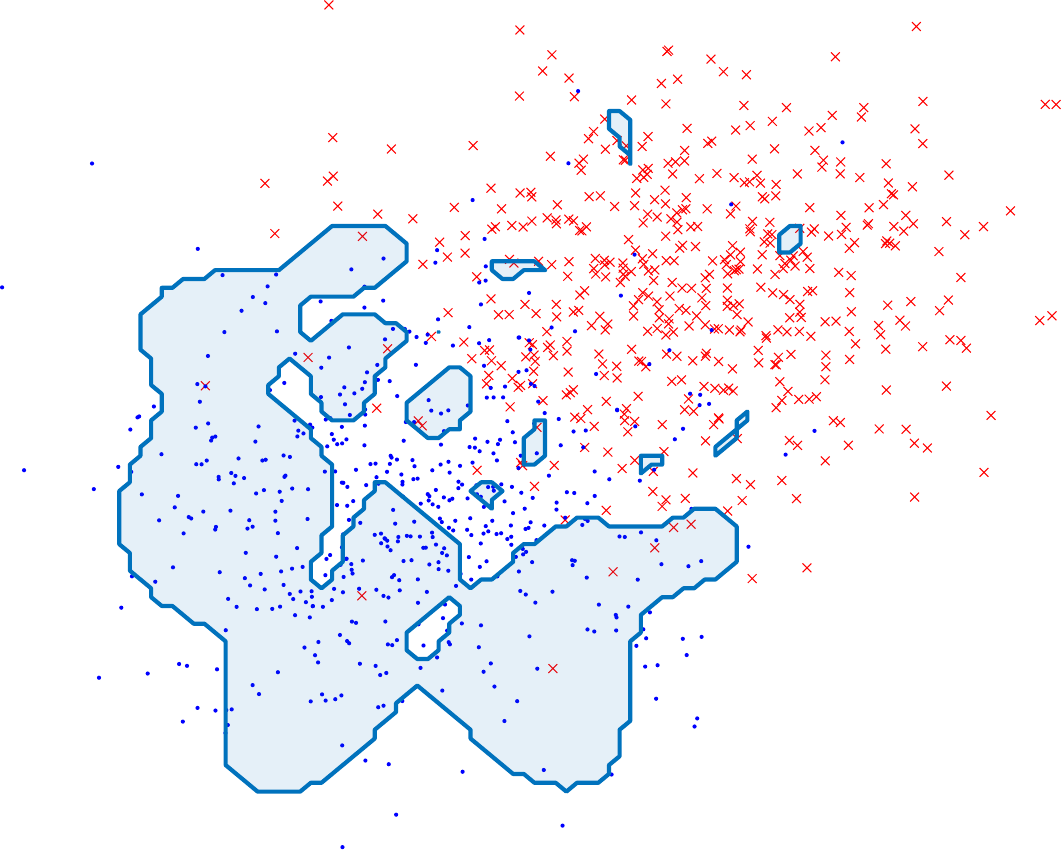}
\label{Fig:FF105}}
\vfil
\subfloat[$\eta = 10^{-2}, \tau = 0.9$ \\ $J(\eta,\tau) = 662$]{\includegraphics[width=2in]{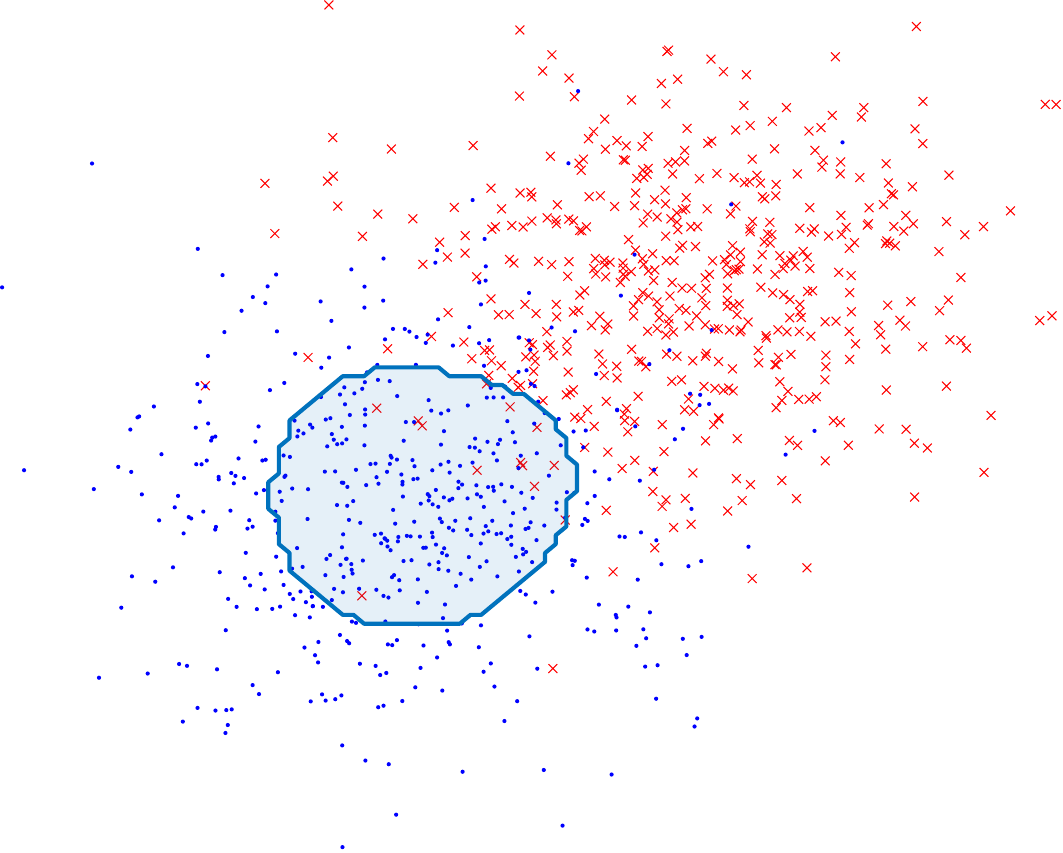}
\label{Fig:FF00109}}
\hfil
\subfloat[$\eta = 10^{-1}, \tau = 0.9$ \\ $J(\eta,\tau) = {\mathbf{811}}$]{\includegraphics[width=2in]{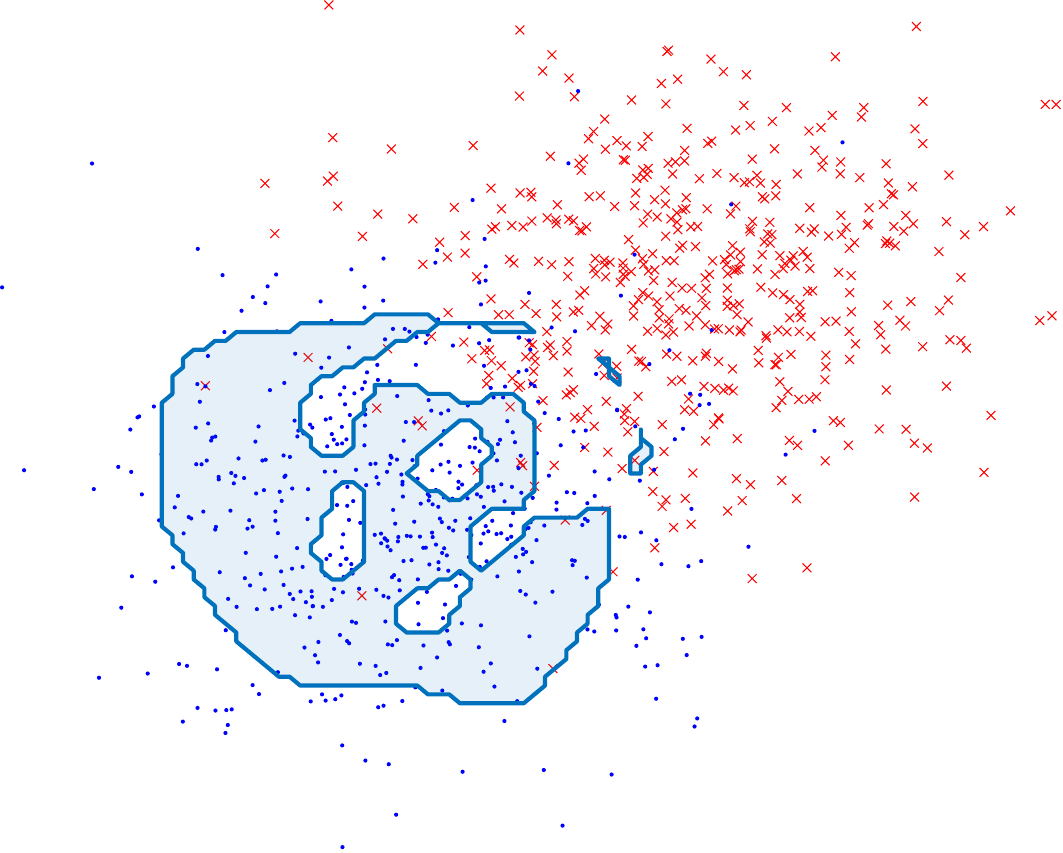}
\label{Fig:FF0109}}
\hfil
\subfloat[$\eta = 1, \tau = 0.9$ \\ $J(\eta,\tau) = 770$]{\includegraphics[width=2in]{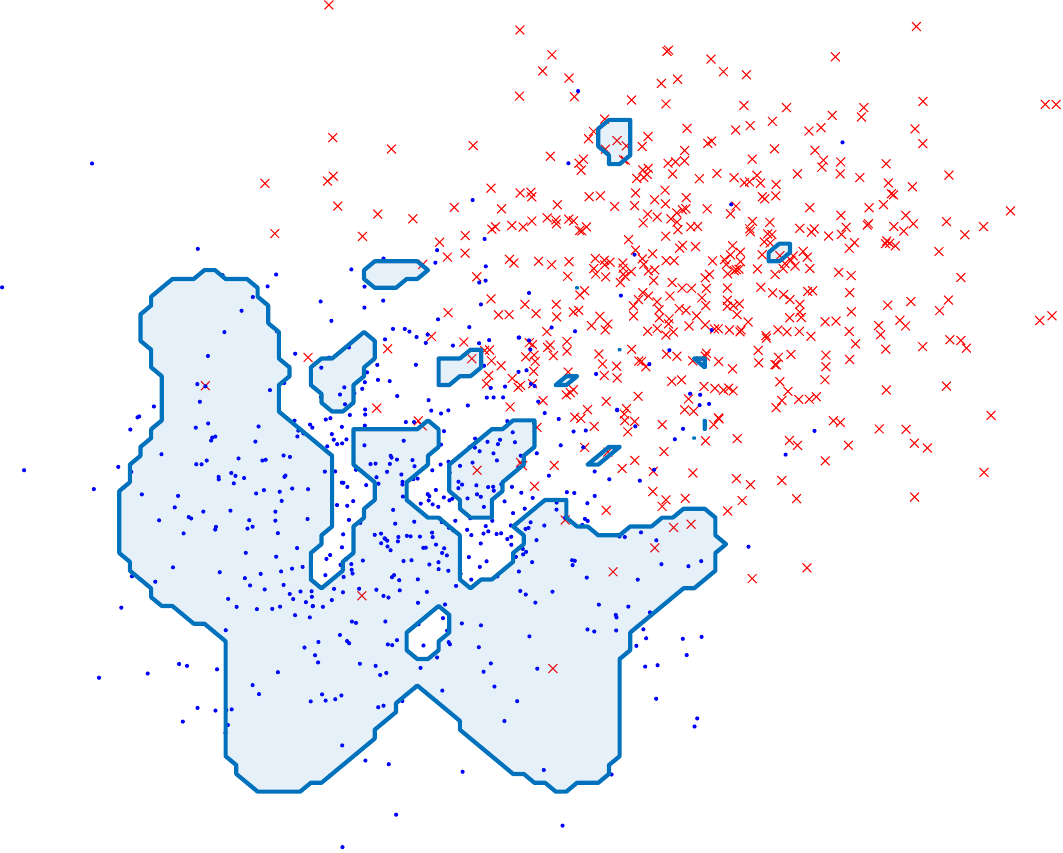}
\label{Fig:FF109}}
\vfil\vskip\baselineskip
\caption{Plots of PSRs at the $varepsilon = 0.05$ level for Gaussian SVDD with different regularization parameters ($\eta$) and different weights ($\tau$). The shape of the region changes by varying the design parameters, but maintaining the probabilistic guarantee on the number of unsafe points within it. The best configuration is chosen by maximizing a performance index, in this case the number of calibration points contained in the region (see the equation \eqref{eq:J function}). For this toy example, the best configuration is obtained for $epsilon = 10^{-1}$ and $tau = 0.9$, but others can be found by increasing the number of candidate design parameters.}
\label{Fig:FF}
\end{figure*}
\begin{theorem}[Probabilistic Safety Region for finite families of hyperparameters]
Consider the classifier \eqref{eq:phi:minus}, a finite set of possible hyperparameter values $\boldsymbol{\theta}\in \Theta=\left\{\thetaB^{(1)}, \thetaB^{(2)}, \dots, \thetaB^{(m)}\right\}$, and suppose that Assumption \ref{assum:Conditions:on:f} holds and that  $\Pr\{\x\in \mathcal{X}\}=1$. Fix a risk parameter $\varepsilon\in (0,1)$, a probability level $\delta\in(0,1)$ and an integer \textit{discarding} parameter $r\ge 1$.
Given $\mathcal{Z}_c^U= \left\{(\tilde{\x}^U_j,-1)\right\}_{j=1}^{n_U} $ corresponding to all the unsafe samples in a calibration set $\mathcal{Z}_c$ of $n_c \ge r$ i.i.d. samples,
for all $\boldsymbol{\theta}^{(k)}$, $k\in[m]$, compute the corresponding scaling factors:
\begin{itemize}
    \item compute the scaling parameters
    $$ \bar\rho_{j}^{(k)} \text{ such that } f_{\thetaB^{(k)}}(\tilde\x_j^U,\bar\rho_{j}^{(k)}) = 0, \quad j \in [n_U],k\in[m],$$
    \item  compute the $k$-th {\em probabilistic radius} and the $k$-th {\em probabilistic safety region of level} $\varepsilon$, i.e.
   \begin{eqnarray}
\rho_\varepsilon^{(k)} &\doteq&
\mathrm{max}^{(r)}\left(\{\bar\rho_{j}^{(k)}\}_{j=1}^{n_U}\right), 
\label{eq:scal_param1}\\
 \mathcal{S}^{(k)}_{\varepsilon} &\doteq& \bsis{cc} \mathcal{S}\left(\rho_\varepsilon^{(k)}\right)  & \text{ if }  n_U\geq r \\ \mathcal{X} & \text{otherwise.} \esis
 \label{eq:scal_param2}. 
\end{eqnarray} 
\end{itemize}
Then, the following holds
\begin{equation}
\label{eq:main_families}
\Pr\left\{\Pr\Bigl\{y=-1 \text{ and } \x \in \mathcal{S}^{(k)}_\varepsilon\Bigr\} \le \varepsilon \right \}\ge 1-m\Bin(r-1;n_c,\varepsilon),
\end{equation}
$\forall k \in [m]$.
\label{theorem:probscalingSRFiniteFamily}
\end{theorem}
In particular, this means that all sets $\mathcal{S}^{(k)}_{\varepsilon}$ are valid PSR candidates, and we have the possibility of selecting among those the ``best" one according to some specific measure on how we expect the SC to behave. In the next subsection, we propose a possible criterion which proved to be very effective in our experience.
\subsection{Increase of safe points}
\label{sec:miss}

In general, one is interested in a solution which, besides providing probabilistic guarantees on the safe region, i.e.\
minimizing the probability of having unsafe points in the set $\mathcal{S}^{(k)}_{\varepsilon}$, it also maximises the number of safe points captured by the region itself.
To this end, we first notice that, when applying the scaling procedure, we are basically only exploiting the unsafe points in 
calibration set $\mathcal{Z}_c$ (i.e.\ the points belonging  to $\mathcal{Z}_c^U$).

It is thus immediately to observe that the remaining points in the calibration set, i.e.\ the points belonging to $$
\mathcal{Z}_c^S=\mathcal{Z}_c\setminus \mathcal{Z}_c^U,
$$
i.e. the set containing all the safe ($+1$)  points in $\mathcal{Z}_c$ may be exploited in evaluating the goodness of the candidate sets.
To this end, given a candidate set $\mathcal{S}^{(k)}_{\varepsilon}$, we can measure its goodness by choosing the cardinality of that set as the performance function 
\begin{equation}
       J(\boldsymbol{\theta}^{(k)}) \doteq \left|\left\{\z\in \mathcal{Z}_c^S: \z\in \mathcal{S}^{(k)}_{\varepsilon}\right\}\right|.
       \label{eq:J function}
\end{equation}
%
\begin{example}
Considering the scalable SVDD with Gaussian kernel, in the same design as Example\ref{ex:GaussianPSR}, but with a probability of sampling outliers per class set at $p_O = 0.1$ (to allow for some noise), with only $1,000$ points for the test set (to make the boundary plot clearer, see Figure \ref{Fig:FF}) and with $\varepsilon$ set to $0.05$ (that gives a calibration set with $2,064$ points), we computed the probabilistic safety region $\mathcal{S}_\varepsilon$ for different values of the hyperparameters $\thetaB = [\eta, \tau]$, specifically $\eta = [10^{-2}, 10^{-1}, 1]$ and $\tau = [0.1,0.2,\dots,0.9]$. All regions satisfy the probabilistic bound on the number of unsafe points within $\mathcal{S}_\varepsilon$, i.e. $\Pr\{\x\lbl\textrm{U}\ \text{and} \ \x\in \mathcal{S}_\varepsilon\}<0.05 $, but the area covered changes as the design parameters change. The best region can be chosen as the one that maximizes an index parameter, as in this case the equation \eqref{eq:J function} that increases the number of safe points in the PSR.

\end{example}

Finally, it is worth noting that, more generally, the parameter to be optimized can be specified in principle according to the specific problem to be solved. For example, $J$ can be defined such that it maximizes accuracy or minimizes only false positives or false negatives or optimizes any other performance index. 
\begin{table*}[htbp]
\label{tab:VPresults}
  \centering
  \caption{Table of the performance of PSRs for vehicle platoon as the collision probability ($\varepsilon = 0.01, 0.05, 0.1$) and design parameters ($\eta = 10^{-2}, 10^{-1}, 1$ and $\tau=0.1, 0.5, 0.9$) variation. The kernel was set as Gaussian. In \textbf{bold}, the \textbf{best} results for each combination of classifier and parameters. }
  \resizebox{1\textwidth}{!}{
    \begin{tabular}{|c|c|r|r|r|r|r|r|r|r|r|r|r|r|r|r|r|r|r|r|}
      \cline{3-20}    
      \multicolumn{1}{r}{} &       & \multicolumn{6}{c|}{$\boldsymbol{\eta=10^{-2}}$} & \multicolumn{6}{c|}{$\boldsymbol{\eta=10^{-1}}$} & \multicolumn{6}{c|}{$\boldsymbol{\eta=1}$} \bigstrut\\
      \cline{3-20}    
      \multicolumn{1}{r}{} &       & \multicolumn{2}{c|}{$\boldsymbol{\tau=0.1}$} & \multicolumn{2}{c|}{$\boldsymbol{\tau=0.5}$} & \multicolumn{2}{c|}{$\boldsymbol{\tau=0.9}$} & \multicolumn{2}{c|}{$\boldsymbol{\tau=0.1}$} & \multicolumn{2}{c|}{$\boldsymbol{\tau=0.5}$} & \multicolumn{2}{c|}{$\boldsymbol{\tau=0.9}$} & \multicolumn{2}{c|}{$\boldsymbol{\tau=0.1}$} & \multicolumn{2}{c|}{$\boldsymbol{\tau=0.5}$} & \multicolumn{2}{c|}{$\boldsymbol{\tau=0.9}$} \bigstrut\\
      \cline{3-20}    
      \multicolumn{1}{r}{} &       & \multicolumn{1}{l|}{$\Pr\{\}$} & \multicolumn{1}{l|}{$J(\boldsymbol{\theta})$} & \multicolumn{1}{l|}{$\Pr\{\}$} & \multicolumn{1}{l|}{$J(\boldsymbol{\theta})$} & \multicolumn{1}{l|}{$\Pr\{\}$} & \multicolumn{1}{l|}{$J(\boldsymbol{\theta})$} & \multicolumn{1}{l|}{$\Pr\{\}$} & \multicolumn{1}{l|}{$J(\boldsymbol{\theta})$} & \multicolumn{1}{l|}{$\Pr\{\}$} & \multicolumn{1}{l|}{$J(\boldsymbol{\theta})$} & \multicolumn{1}{l|}{$\Pr\{\}$} & \multicolumn{1}{l|}{$J(\boldsymbol{\theta})$} & \multicolumn{1}{l|}{$\Pr\{\}$} & \multicolumn{1}{l|}{$J(\boldsymbol{\theta})$} & \multicolumn{1}{l|}{$\Pr\{\}$} & \multicolumn{1}{l|}{$J(\boldsymbol{\theta})$} & \multicolumn{1}{l|}{$\Pr\{\}$} & \multicolumn{1}{l|}{$J(\boldsymbol{\theta})$} \bigstrut\\
      \cline{3-20}    
      \multicolumn{1}{r}{} &       & \multicolumn{18}{c|}{$\boldsymbol{\varepsilon = 0.01}$} \bigstrut\\
      \hline
      \multirow{11}[22]{*}{\rotatebox[origin=c]{90}{\large \textsc{Vehicle Platooning}}} & \textsc{SC-SVM} & \multicolumn{1}{c|}{0.04} & \multicolumn{1}{c|}{825} & \multicolumn{1}{c|}{0.02} & \multicolumn{1}{c|}{1254} & \multicolumn{1}{c|}{0.03} & \multicolumn{1}{c|}{1054} & \multicolumn{1}{c|}{0.02} & \multicolumn{1}{c|}{1211} & \multicolumn{1}{c|}{0.01} & \multicolumn{1}{c|}{1839} & \multicolumn{1}{c|}{0.01} & \multicolumn{1}{c|}{2030} & \multicolumn{1}{c|}{0.01} & \multicolumn{1}{c|}{1390} & \multicolumn{1}{c|}{0.01} & \multicolumn{1}{c|}{1833} & \multicolumn{1}{c|}{${\textbf{0.01}}$} & \multicolumn{1}{c|}{${\textbf{2361}}$} \bigstrut\\
      \cline{2-20}          
      & \textsc{SC-SVDD} & \multicolumn{1}{c|}{0.01} & \multicolumn{1}{c|}{3347} & \multicolumn{1}{c|}{0.02} & \multicolumn{1}{c|}{2395} & \multicolumn{1}{c|}{0.01} & \multicolumn{1}{c|}{3532} & \multicolumn{1}{c|}{0.01} & \multicolumn{1}{c|}{3342} & \multicolumn{1}{c|}{0.02} & \multicolumn{1}{c|}{2395} & \multicolumn{1}{c|}{0.01} & \multicolumn{1}{c|}{3547} & \multicolumn{1}{c|}{0.01} & \multicolumn{1}{c|}{3342} & \multicolumn{1}{c|}{0.01} & \multicolumn{1}{c|}{3032} & \multicolumn{1}{c|}{{\textbf{0.01}}} & \multicolumn{1}{c|}{{\textbf{3548}}} \bigstrut\\
      \cline{2-20}          
      & \textsc{SC-LR}  & \multicolumn{1}{c|}{0.02} & \multicolumn{1}{c|}{5373} & \multicolumn{1}{c|}{0.02} & \multicolumn{1}{c|}{5372} & \multicolumn{1}{c|}{0.02} & \multicolumn{1}{c|}{5372} & \multicolumn{1}{c|}{{\textbf{0.02}}}& \multicolumn{1}{c|}{{\textbf{5378}}} & \multicolumn{1}{c|}{0.02} & \multicolumn{1}{c|}{5373} & \multicolumn{1}{c|}{0.02} & \multicolumn{1}{c|}{5372} & \multicolumn{1}{c|}{0.02} & \multicolumn{1}{c|}{5372} & \multicolumn{1}{c|}{0.02} & \multicolumn{1}{c|}{5376} & \multicolumn{1}{c|}{0.02} & \multicolumn{1}{c|}{5350} \bigstrut\\
      \cline{2-20}          
      &       & \multicolumn{18}{c|}{$\boldsymbol{\varepsilon = 0.05}$} \bigstrut\\
      \cline{2-20}          
      & \textsc{SC-SVM} & \multicolumn{1}{c|}{0.06} & \multicolumn{1}{c|}{486} & \multicolumn{1}{c|}{0.07} & \multicolumn{1}{c|}{568} & \multicolumn{1}{c|}{0.08} & \multicolumn{1}{c|}{518} & \multicolumn{1}{c|}{0.05} & \multicolumn{1}{c|}{707} & \multicolumn{1}{c|}{0.05} & \multicolumn{1}{c|}{780} & \multicolumn{1}{c|}{0.06} & \multicolumn{1}{c|}{752} & \multicolumn{1}{c|}{0.05} & \multicolumn{1}{c|}{805} & \multicolumn{1}{c|}{0.04} & \multicolumn{1}{c|}{797} & \multicolumn{1}{c|}{{\textbf{0.05}}} & \multicolumn{1}{c|}{{\textbf{844}}} \bigstrut\\
      \cline{2-20}          
      & \textsc{SC-SVDD} & \multicolumn{1}{c|}{0.05} & \multicolumn{1}{c|}{886} & \multicolumn{1}{c|}{0.06} & \multicolumn{1}{c|}{678} & \multicolumn{1}{c|}{0.05} & \multicolumn{1}{c|}{876} & \multicolumn{1}{c|}{{\textbf{0.05}}} & \multicolumn{1}{c|}{{\textbf{889}}} & \multicolumn{1}{c|}{0.06} & \multicolumn{1}{c|}{678} & \multicolumn{1}{c|}{0.05} & \multicolumn{1}{c|}{880} & \multicolumn{1}{c|}{{\textbf{0.05}}} & \multicolumn{1}{c|}{{\textbf{889}}} & \multicolumn{1}{c|}{0.06} & \multicolumn{1}{c|}{763} & \multicolumn{1}{c|}{0.05} & \multicolumn{1}{c|}{880} \bigstrut\\
      \cline{2-20}          
      & \textsc{SC-LR}  & \multicolumn{1}{c|}{0.03} & \multicolumn{1}{c|}{909} & \multicolumn{1}{c|}{0.03} & \multicolumn{1}{c|}{910} & \multicolumn{1}{c|}{0.03} & \multicolumn{1}{c|}{909} & \multicolumn{1}{c|}{0.03} & \multicolumn{1}{c|}{915} & \multicolumn{1}{c|}{0.03} & \multicolumn{1}{c|}{911} & \multicolumn{1}{c|}{0.02} & \multicolumn{1}{c|}{907} & \multicolumn{1}{c|}{{\textbf{0.03}}} & \multicolumn{1}{c|}{{\textbf{953}}} & \multicolumn{1}{c|}{0.03} & \multicolumn{1}{c|}{951} & \multicolumn{1}{c|}{0.00} & \multicolumn{1}{c|}{889}\bigstrut\\
      \cline{2-20}          
      &       &\multicolumn{18}{c|}{$\boldsymbol{\varepsilon = 0.1}$} \bigstrut\\
      \cline{2-20}          
      & \textsc{SC-SVM} & \multicolumn{1}{c|}{0.10} & \multicolumn{1}{c|}{360} & \multicolumn{1}{c|}{0.12} & \multicolumn{1}{c|}{394} & \multicolumn{1}{c|}{0.16} & \multicolumn{1}{c|}{357} & \multicolumn{1}{c|}{0.14} & \multicolumn{1}{c|}{430} & \multicolumn{1}{c|}{0.10} & \multicolumn{1}{c|}{456} & \multicolumn{1}{c|}{0.10} & \multicolumn{1}{c|}{449} & \multicolumn{1}{c|}{0.12} & \multicolumn{1}{c|}{466} & \multicolumn{1}{c|}{0.09} & \multicolumn{1}{c|}{495} & \multicolumn{1}{c|}{{\textbf{0.09}}} & \multicolumn{1}{c|}{{\textbf{487}}} \bigstrut\\
      \cline{2-20}          
      & \textsc{SC-SVDD} & \multicolumn{1}{c|}{0.10} & \multicolumn{1}{c|}{508} & \multicolumn{1}{c|}{0.11} & \multicolumn{1}{c|}{463} & \multicolumn{1}{c|}{0.09} & \multicolumn{1}{c|}{528} & \multicolumn{1}{c|}{0.10} & \multicolumn{1}{c|}{508} & \multicolumn{1}{c|}{0.11} & \multicolumn{1}{c|}{463} & \multicolumn{1}{c|}{{\textbf{0.09}}} & \multicolumn{1}{c|}{{\textbf{529}}} & \multicolumn{1}{c|}{0.09} & \multicolumn{1}{c|}{508} & \multicolumn{1}{c|}{0.10} & \multicolumn{1}{c|}{485} & \multicolumn{1}{c|}{{\textbf{0.09}}} & \multicolumn{1}{c|}{{\textbf{529}}} \bigstrut\\
      \cline{2-20}          
      & \textsc{SC-LR}  & \multicolumn{1}{c|}{0.10} & \multicolumn{1}{c|}{566} & \multicolumn{1}{c|}{0.10} & \multicolumn{1}{c|}{574} & \multicolumn{1}{c|}{0.10} & \multicolumn{1}{c|}{574} & \multicolumn{1}{c|}{0.10} & \multicolumn{1}{c|}{577} & \multicolumn{1}{c|}{0.09} & \multicolumn{1}{c|}{586} & \multicolumn{1}{c|}{0.09} & \multicolumn{1}{c|}{567} & \multicolumn{1}{c|}{{\textbf{0.09}}} & \multicolumn{1}{c|}{{\textbf{597}}} & \multicolumn{1}{c|}{0.07} & \multicolumn{1}{c|}{568} & \multicolumn{1}{c|}{0.06} & \multicolumn{1}{c|}{558}   \bigstrut\\
      \hline
    \end{tabular}
  }
  \label{tab:addlabel}
\end{table*}
\section{A real-world application:\\ Vehicle Platooning}

Safety critical assessment represents a fundamental requirement in the automotive industry and vehicle platooning (VP) \cite{7056505} represents one of the most challenging CPS (Cyber Physical System) in this context. The main goal of VP is to find the best trade-off between performance (i.e., maximizing speed and minimizing vehicle mutual distance) and safety (i.e., collision avoidance). With the idea of finding the largest region in the input space where safety is probabilistically guaranteed, we tested our scalable classifiers on the following scenario: given the platoon at a steady state of speed and reciprocal distance of the vehicles, a braking is applied by the leader of the platoon \cite{Xu,Santini}. Safety is referred to a collision between adjacent vehicles
(in the study, it is actually registered when the reciprocal distance between vehicles achieves a lower bound, e.g. 2 m). 
The dynamic of the system is generated by the following differential equations \cite{Xu}:

\begin{equation}
    \begin{cases}
    \dot{v_\ell}=\frac{1}{m_\ell}(F_\ell-(a_\ell+b_\ell\cdot v_\ell^2)) \\
    \dot{d}_\ell=v_{\ell-1}-v_\ell
    \end{cases}
    \label{29}
\end{equation}

\noindent where $v_\ell, m_\ell, a_\ell, b_\ell$ and $F_\ell$ are, respectively, the speed, the mass, the tire-road rolling distance, the aerodynamic drag and the braking force (the control law) of vehicle $\ell$ and $d_\ell$ is the distance of vehicle $\ell$ from the previous one $\ell-1$. 

The behaviour of the dynamical system is synthesised by the
following vector of features:
\begin{equation}
    \x=[N, \boldsymbol{\iota}(0), F_0, \mathbf{m}, \mathbf{q}, \mathbf{p}],
    \label{30}
\end{equation}

\noindent $N + 1$ being the number of vehicles in the platoon, $\boldsymbol{\iota}=[\mathbf{d}, \mathbf{v}, \mathbf{a}]$ are the vectors of reciprocal distance, speed, and acceleration of the vehicles, respectively ($\boldsymbol{\iota}$(0) denotes that the quantities are sampled at time $t=0$, after which a braking force is applied by the leader \cite{Santini} and simulations are set in order to manage possible transient periods and achieve a steady state of $\boldsymbol{\iota}$ before applying the braking), \textbf{m} is the vector of
weights of the vehicles, $F_0$ is the braking force applied by the leader, \textbf{q} is the vector of quality measures of the communication medium (fixed delay and packet error rate (PER) are considered in the simulations) and finally \textbf{p} is the vector of tuning parameters of the control scheme.

The Plexe simulator \cite{Santini,Xu} has been used to register $20000$ observations in the following ranges:
\noindent $N\in[3,8], \ F_0\in[-8,-1]\times10^3 N$, $\mathbf{q}\in [0,0.5], \ \mathbf{d}(0)\in [4,9] \ \text{m}, \ \mathbf{v}(0)\in[10,90] \ \text{Km/h}$. Initial acceleration $\mathbf{a}(0)$ is computed as $\mathbf{a}(0) = F_0/\mathbf{m}\ \text{Km/h}^2$. The output variable is then defined as $y\in\mathcal{Y} = \{-1,+1\}$, where $-1$ means ``collision'' and $+1$ means ``non-collision''.\\
We searched safety for three levels of guarantee ($\varepsilon = 0.01, 0.05, 0.1$) and different hyperparameters ($\eta = 10^{-2},10^{-1},1$ and $\tau = 0.1, 0.5, 0.9$), evaluating the performance on the test set (reported in Table \ref{tab:VPresults}) computing the empirical probability of getting a collision inside the ``non-collision'' probabilistic safety region $\mathcal{S}_\varepsilon$, $\Pr\Bigl\{\x\lbl\{\text{collision}\} \ \text{and} \ \mathbf{\x}\in\mathcal{S}_\varepsilon\Bigr\}$, and the number of non-collision points of the calibration set contained in $\mathcal{S}_\varepsilon$, varying the hyperparameters, $J(\boldsymbol{\eta}, \boldsymbol{\tau})$.

\begin{figure}[!h]
\centering
\includegraphics[width=0.42\textwidth]{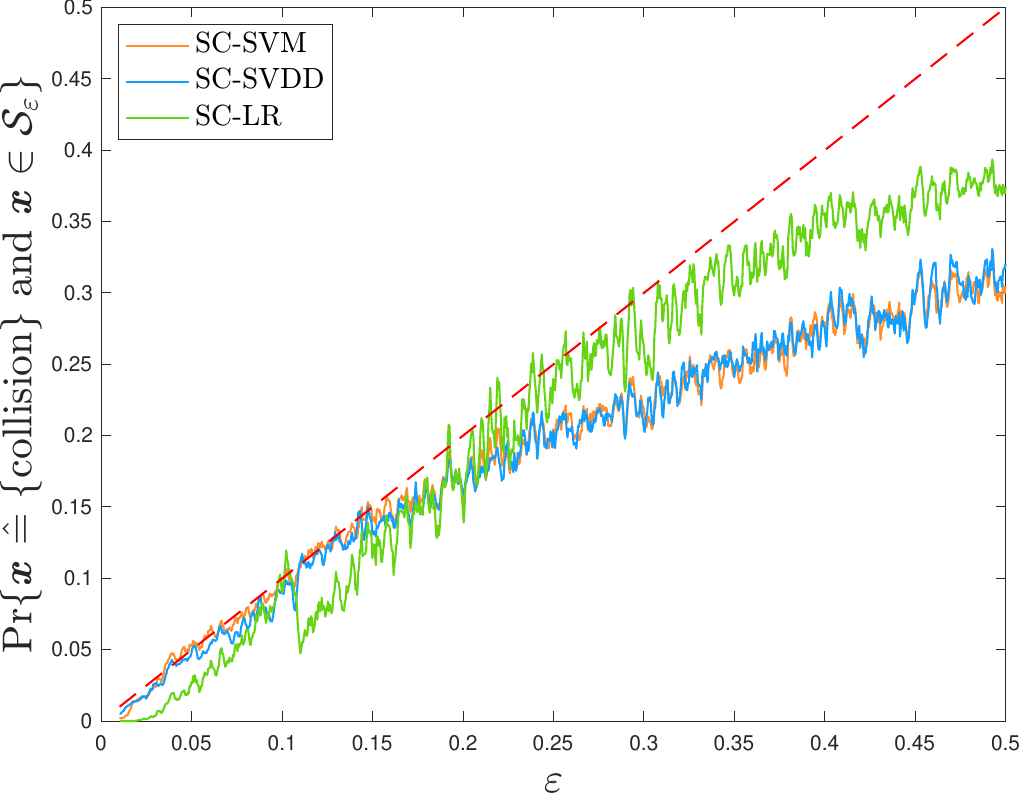}
\caption{Empirical probability of collision, measured in the test set, as a function of $\varepsilon$.} 
\label{Fig:plotEPS}
\end{figure}

We divided the dataset in training set ($n_{tr} = 3000$ points), calibration set ($n_c = 10320, 2064, 1032$ 
respectively for $\varepsilon = 0.01, 0.05,0.1$) and test set ($n_{ts} = n_{tr} - n_c$). \\
In this numerical example, we use the scalable classifiers presented in Subsection \ref{Sub:sec:Examples}; scalable SVM (SC-SVM), scalable SVDD (SC-SVDD) and scalable LR (SC-LR). In all of them, the Gaussian kernel has been employed. 
For all scalable classifiers, the trade-off between the guarantee and the number of safe points of the calibration set within the ``non-collision'' safety region is good, allowing for the construction of operational regions where safety can be guaranteed. In particular, the best performance obtained by each classifier at different levels of $\varepsilon$ is highlighted in bold in Table \ref{tab:VPresults}.
 Furthermore, Figure \ref{Fig:plotEPS} shows the trend of the probability of getting a collision within the safety region as $\varepsilon$ varies, with $\eta = 1$ and $\tau = 0.5$ (i.e., without regularizing and weighting equally both the classes). As expected, the behavior is (almost) linear with $\varepsilon$, with SC-LR deviating slightly from SC-SVM and SC-SVDD.\\
It should be mentioned that this work represents a significant improvement of the results obtained on the same dataset by \cite{9594676}. In this previous research, a safety set was searched by numerically minimizing the number of false positives by controlling the radius of an SVDD classifier. Here, with the new theory based on SCs and probabilistic scaling, we obtain better results in terms of performance (i.e., size of safety regions) and, more importantly,  with a \textit{formally solid mathematical framework}, applicable to any binary classifier.

\section{Conclusions}
\label{sec:conclusions}

Statistical learning can rely on two significant new concepts: scalable classifiers and probabilistic safety regions 
SCs constitute a new family of classifiers, expanding the knowledge in the field of ML classification, which has been state of the art for many years. Three examples of SCs have been proposed (SC-SVM, SC-SVDD and SC-LR), but more can be developed. For example, the results of the paper can be straightforwardly applied in the field of deep neural networks, which, in principle, would have a crucial impact: they would allow classification performance to be controlled without lengthy retraining. Also, the definition of PSR and its \underline{proved} properties provide the field of CPSs with a new relevant methodology for evaluating safety. The range of applications in which PSRs can be introduced is definitely wide: from safety monitoring to performance improvement and conformity guarantee, just to name a few. Moreover, the ideas here introduced open up new research directions and parallels with other theories. In particular, and with this remark we conclude the paper, our theory has much in common with Conformal Prediction, allowing for connections that both methodologies can use to improve each other.

\begin{remark}[Link with Conformal Prediction]
\label{remark:CP}
We shall remark that our approach bears various similarities with the Conformal Prediction approach.
    Conformal Prediction is a framework developed starting in the late nineties and early two thousand by V.\ Vovk. We refer the reader to the surveys \cite{gentleIntro,tutorial_cP,Survey_cP} for a nice introduction to this methodology. 
    Similarly to our approach, CP makes use of a calibration set to perform an a-posteriori verification of the designed classifier. In practice, it returns a measure of its "conformity" to the calibration data, based on the introduction of specific score functions.
We are currently working in showing how specific score functions may be designed in the case of scalable classifier. Preliminary results, which go beyond the scope of present work, seem to prove how the scaling setup proposed here may be used to prove interesting safety properties of CPs.
\end{remark}

\section*{Acknowledgment}
We would like to thank \href{https://people.eecs.berkeley.edu/~angelopoulos/}{A.\ Angelopoulos} for the inspiring discussions on possible links between scalable classifiers and conformal predictions.

\appendix

This appendix is devoted to the proofs and derivations of all the original results presented in this paper.

\label{sec:proofs}
\subsection{Property \ref{prop:existence:uniqueness}}
\label{proof: property1}
\begin{proof}
Because of \eqref{eq:limits:f:prop} we have that if $\rho$ is small enough then $f_\thB(\x,\rho)<0$. On the other hand, if $\rho$ is large enough then 
$f_\thB(\x,\rho)>0$. This, along with the continuity nature of $f_\thB(\x,\rho)$, guarantees the existence of $\rho$ such that $f_\thB(\x,\rho)=0$. 
The uniqueness follows from the monotonic assumption on $f_\thB(\x,\rho)$. Denote $\bar{\rho}(\x)$ this unique value of $\rho$ satisfying  $f_\thB(\x,\bar{\rho}(\x))=0$. 
From the monotonically increasing nature of $f_\thB(\x,\rho)$ we have
$$ f_\thB(\x,\rho) \geq 0 \;\; \Leftrightarrow \; \; \rho \geq \bar{\rho}(\x).$$
Thus, 
$$ \phi_\theta(\x,\rho) = -1 \Leftrightarrow \rho \geq \bar{\rho}(\x).$$
\end{proof}

\subsection{Theorem \ref{theorem:main}}
\label{proof:teo-main}
\begin{proof} 
Let us introduce the auxiliary function $$\psi:\mathcal{X}\times\{-1,1\}\to [-1,1),$$ which is defined as 
\begin{equation} \label{eq:tau:def}
     \psi(\x,y)  \doteq \bsis{cl}
        -1 & \text{if } \, y =+1, \\
        \frac{\bar{\rho}(\x)}{1+|\bar{\rho}(\x)|} &\text{otherwise.}
    \esis
\end{equation}
Denote now $$\psi_\varepsilon = \mathrm{max}^{(r)}\left(\{\psi(\x_i,y_i)\}_{i=1}^{n_c}\right) ).$$
Since $\delta\in (0,1)$, $\varepsilon\in (0,1)$ and the integers $n_c\geq r\geq 1$ satisfy
$$  \Sum{i=0}{r-1}\binom{n_c}{i}\varepsilon^i(1-\varepsilon)^{n_c-i}\leq \delta,$$
we have from Property \ref{Property:Generalized:Max} that, with a probability no smaller than $1-\delta$, 
\begin{equation}\label{ineq:tau:epsilon}
    \Pr\Bigl\{ \psi(\x,y) > \psi_\epsilon \Bigr\} \leq \varepsilon.
\end{equation}
The rest of the proof shows that the previous inequality is equivalent to the claim of the theorem. 
That is, 
$$ \Pr\Bigl\{ \psi(\x,y) > \psi_\varepsilon \Bigr\} \leq \varepsilon \; \; \Leftrightarrow \; \; \Prob\Bigl\{y=-1 \text{ and } \x \in S_\epsilon\Bigr\}  \leq \varepsilon.$$
We consider two cases $n_U<r$ and $n_U\geq r$.
\begin{itemize}
    \item {\bf Case $n_U<r$}:  By definition, 
    $$ -1=\psi(\x,+1) < \psi(\x,-1)\in(-1,1), \; \forall \x\in \mathcal{X}.$$
    This means that the smallest values for $\psi(\x,y)$ are attained at the safe samples. 
    From $n_U<r$ we have that at most $r-1$ elements of the calibration set correspond to unsafe samples. Equivalently, no more than $r-1$ elements of $\{\psi(\x_i,y_i)\}_{i=1}^{n_c}$ are larger than $-1$. This implies that the $r$-th largest value in 
    $\{\psi(\x_i,y_i)\}_{i=1}^{n_c}$ corresponds to a safe sample and is equal to $-1$. That is, 
     $$ \psi_\varepsilon = \mathrm{max}^{(r)}\left(\{\psi(\x_i,y_i)\}_{i=1}^{n_c}\right) ) = -1. $$
     Thus, the inequality \eqref{ineq:tau:epsilon} is equivalent in this case to 
     $$ \Pr\Bigl\{ \psi(\x,y) > -1 \Bigr\} \leq \varepsilon.$$ 
     By definition, for every $\x\in \mathcal{X}$ we have $$\psi(\x,y) >-1 \; \Leftrightarrow\; y=-1. $$ 
Thus, we obtain that in this case, $\psi_\varepsilon=-1$ and 
\begin{eqnarray*}
\Pr\Bigl\{ \psi(\x,y) > \psi_\varepsilon \Bigr\} \leq \varepsilon & \Leftrightarrow & \Pr\Bigl\{ y=-1 \Bigr\} \leq \varepsilon \\
& \Leftrightarrow & \Pr\Bigl\{ y=-1  \text{ and } \x\in \mathcal{X}\Bigr\} \leq \varepsilon.
\end{eqnarray*}
   From the assumptions of the Theorem, we have that, by definition, $n_U < r$ implies $S_\varepsilon= \mathcal{X}$. Thus, we  conclude that in this case, 
   $$ \Pr\Bigl\{ \psi(\x,y) > \psi_\varepsilon \Bigr\} \leq \varepsilon \; \Leftrightarrow \Pr\Bigl\{ y=-1 \text{ and } \x\in S_\varepsilon \Bigr\} \leq \varepsilon. $$

    \item {\bf Case $n_U\geq r$}: In this case, the $r$-largest value of $ \{\psi(\x_i,y_i)\}_{i=1}^{n_c} $ is attained at an element of the unsafe calibration set $\mathcal{Z}_c^U= \left\{(\tilde{\x}^U_j,-1)\right\}_{j=1}^{n_U} \subseteq \mathcal{Z}_c$. That is, 
  \begin{eqnarray*} 
  \psi_\varepsilon &=& \mathrm{max}^{(r)}\left(\{\psi(\x_i,y_i)\}_{i=1}^{n_c}\right)  \\
  &=& \mathrm{max}^{(r)} (\{\psi(\tilde{\x}^U_j,-1)\}_{j=1}^{n_U})\in (-1,1).
  \end{eqnarray*}
Define now 
$$\rho_\varepsilon = \mathrm{max}^{(r)}\left(\{\bar{\rho}(\tilde{\x}^U_{j})\}_{j=1}^{n_U}\right).$$
Since  $\frac{\bar{\rho}(\x)}{1+|\bar{\rho}(\x)|}$ is a monotonically increasing function on $\bar{\rho}(\x)$, we have that $\psi_\epsilon$ can be obtained by means of $\rho_\varepsilon$. That is, 
$$\psi_\varepsilon = \mathrm{max}^{(r)} (\{\psi(\tilde{\x}^U_j,-1)\}_{j=1}^{n_U})= \frac{\rho_\varepsilon}{1+|\rho_\varepsilon|}.$$
Thus, from $\psi_\varepsilon>-1$ and the previous expression we obtain the equivalences
 \begin{eqnarray*} \psi(\x,y) > \psi_\varepsilon &\Leftrightarrow &  y=-1 \text{ and } \frac{\bar{\rho}(\x)}{1+|\bar{\rho}(\x)|} > \frac{\rho_\varepsilon}{1+|\rho_\varepsilon|} \\
 &\Leftrightarrow & y=-1 \text{ and } \bar{\rho}(\x) \geq \rho_\varepsilon.
 \end{eqnarray*}
 Therefore, $ \Pr\Bigl\{ \psi(x,y) > \psi_\varepsilon \Bigr\} \leq \varepsilon$ is equivalent to 
 $$ \Pr\Bigl\{ y=-1 \text{ and } \bar{\rho}(\x) > \rho_\varepsilon \Bigr\} \leq \varepsilon. $$
From the monotonicity of $f_\theta(\x,\rho)$ on $\rho$ (Assumption \ref{assum:Conditions:on:f}) we obtain that the previous inequality can be rewritten as 
$$ \Pr\Bigl\{ y=-1 \text{ and } f_\theta(\x,\bar{\rho}(x))  > f_\theta(\x,  \rho_\varepsilon ) \Bigr\} \leq \varepsilon.$$
Taking into consideration that $f_\theta(\x,\bar{\rho}(x) )=0$ we obtain that  
$ \Pr\Bigl\{ \psi(x,y) > \psi_\varepsilon \Bigr\} \leq \varepsilon$ is equivalent to 
$$ \Pr\Bigl\{ y=-1 \text{ and } f_\theta(\x, \rho_\varepsilon) <0 \Bigr\} \leq \varepsilon. $$
From the assumptions of the Theorem we have that, by definition, $n_U\geq r$ implies that $S_\varepsilon$ is equal to $\set{\x\in \mathcal{X}}{f_\theta(\x, \rho_\varepsilon) <0 } $. Thus, we conclude in this case that  
$$ \Pr\Bigl\{ \psi(\x,y) > \psi_\varepsilon \Bigr\} \leq \varepsilon \; \Leftrightarrow \Pr\Bigl\{ y=-1 \text{ and } \x\in S_\varepsilon \Bigr\} \leq \varepsilon. $$
\end{itemize}

\end{proof}


\ifCLASSOPTIONcaptionsoff
  \newpage
\fi

\bibliographystyle{IEEEtran}
\bibliography{biblio.bib}

\begin{thebibliography}{10}
\providecommand{\url}[1]{#1}
\csname url@samestyle\endcsname
\providecommand{\newblock}{\relax}
\providecommand{\bibinfo}[2]{#2}
\providecommand{\BIBentrySTDinterwordspacing}{\spaceskip=0pt\relax}
\providecommand{\BIBentryALTinterwordstretchfactor}{4}
\providecommand{\BIBentryALTinterwordspacing}{\spaceskip=\fontdimen2\font plus
\BIBentryALTinterwordstretchfactor\fontdimen3\font minus
  \fontdimen4\font\relax}
\providecommand{\BIBforeignlanguage}[2]{{%
\expandafter\ifx\csname l@#1\endcsname\relax
\typeout{** WARNING: IEEEtran.bst: No hyphenation pattern has been}%
\typeout{** loaded for the language `#1'. Using the pattern for}%
\typeout{** the default language instead.}%
\else
\language=\csname l@#1\endcsname
\fi
#2}}
\providecommand{\BIBdecl}{\relax}
\BIBdecl

\bibitem{Amodei}
\BIBentryALTinterwordspacing
D.~Amodei, C.~Olah, J.~Steinhardt, P.~F. Christiano, J.~Schulman, and
  D.~Man{\'{e}}, ``Concrete problems in {AI} safety,'' \emph{CoRR}, vol.
  abs/1606.06565, 2016. [Online]. Available:
  \url{http://arxiv.org/abs/1606.06565}
\BIBentrySTDinterwordspacing

\bibitem{9074237}
S.~Jain, M.~Luthra, S.~Sharma, and M.~Fatima, ``Trustworthiness of artificial
  intelligence,'' in \emph{2020 6th International Conference on Advanced
  Computing and Communication Systems (ICACCS)}, 2020, pp. 907--912.

\bibitem{doi:10.2514/6.2020-1521}
\BIBentryALTinterwordspacing
T.~J. Wignall and H.~Houlden, \emph{Uncertainty Quantification for Launch
  Vehicle Aerodynamic Lineloads}. [Online]. Available:
  \url{https://arc.aiaa.org/doi/abs/10.2514/6.2020-1521}
\BIBentrySTDinterwordspacing

\bibitem{RePEc:ofr:wpaper:15-19}
\BIBentryALTinterwordspacing
J.~Chen, M.~D. Flood, and R.~B. Sowers, ``{Measuring the Unmeasurable: An
  Application of Uncertainty Quantification to Financial Portfolios},'' Office
  of Financial Research, US Department of the Treasury, Working Papers 15-19,
  Oct. 2015. [Online]. Available:
  \url{https://ideas.repec.org/p/ofr/wpaper/15-19.html}
\BIBentrySTDinterwordspacing

\bibitem{zou2023review}
K.~Zou, Z.~Chen, X.~Yuan, X.~Shen, M.~Wang, and H.~Fu, ``A review of
  uncertainty estimation and its application in medical imaging,'' 2023.

\bibitem{safeautonomousdriving}
B.~Xu, Q.~Li, T.~Guo, Y.~Ao, and D.~Du, ``A quantitative safety verification
  approach for the decision-making process of autonomous driving,'' in
  \emph{2019 International Symposium on Theoretical Aspects of Software
  Engineering (TASE)}, 2019, pp. 128--135.

\bibitem{Jorgensen_2023}
\BIBentryALTinterwordspacing
S.~Jorgensen, J.~Holodnak, J.~Dempsey, K.~de~Souza, A.~Raghunath, V.~Rivet,
  N.~DeMoes, A.~Alejos, and A.~Wollaber, ``Extensible machine learning for
  encrypted network traffic application labeling via uncertainty
  quantification,'' \emph{{IEEE} Transactions on Artificial Intelligence}, pp.
  1--15, 2023. [Online]. Available:
  \url{https://doi.org/10.1109%2Ftai.2023.3244168}
\BIBentrySTDinterwordspacing

\bibitem{BoLi}
\BIBentryALTinterwordspacing
B.~Li, P.~Qi, B.~Liu, S.~Di, J.~Liu, J.~Pei, J.~Yi, and B.~Zhou, ``Trustworthy
  {AI:} from principles to practices,'' \emph{CoRR}, vol. abs/2110.01167, 2021.
  [Online]. Available: \url{https://arxiv.org/abs/2110.01167}
\BIBentrySTDinterwordspacing

\bibitem{BruceGuardrail}
B.~Nagy, ``System safety considerations for al/ml (categorization/attributes)
  versus traditional code regarding interlocks.''\hskip 1em plus 0.5em minus
  0.4em\relax Naval Ordinance Safety and Security Activity (NOSSA) workshop,
  2023.

\bibitem{LUCAS20084591}
\BIBentryALTinterwordspacing
L.~Lucas, H.~Owhadi, and M.~Ortiz, ``Rigorous verification, validation,
  uncertainty quantification and certification through concentration-of-measure
  inequalities,'' \emph{Computer Methods in Applied Mechanics and Engineering},
  vol. 197, no.~51, pp. 4591--4609, 2008. [Online]. Available:
  \url{https://www.sciencedirect.com/science/article/pii/S0045782508002326}
\BIBentrySTDinterwordspacing

\bibitem{8595071}
R.~Yan, J.~Yang, D.~Zhu, and K.~Huang, ``Design verification and validation for
  reliable safety-critical autonomous control systems,'' in \emph{2018 23rd
  International Conference on Engineering of Complex Computer Systems
  (ICECCS)}, 2018, pp. 170--179.

\bibitem{David2011}
\BIBentryALTinterwordspacing
H.~A. David, \emph{Order Statistics}.\hskip 1em plus 0.5em minus 0.4em\relax
  Berlin, Heidelberg: Springer Berlin Heidelberg, 2011, pp. 1039--1040.
  [Online]. Available: \url{https://doi.org/10.1007/978-3-642-04898-2_436}
\BIBentrySTDinterwordspacing

\bibitem{MAMMARELLA2022110108}
\BIBentryALTinterwordspacing
M.~Mammarella, V.~Mirasierra, M.~Lorenzen, T.~Alamo, and F.~Dabbene,
  ``Chance-constrained sets approximation: A probabilistic scaling approach,''
  \emph{Automatica}, vol. 137, p. 110108, 2022. [Online]. Available:
  \url{https://www.sciencedirect.com/science/article/pii/S0005109821006373}
\BIBentrySTDinterwordspacing

\bibitem{Alamo2018}
\BIBentryALTinterwordspacing
T.~Alamo, J.~M. Manzano, and E.~F. Camacho, \emph{Robust Design Through
  Probabilistic Maximization}.\hskip 1em plus 0.5em minus 0.4em\relax Cham:
  Springer International Publishing, 2018, pp. 247--274. [Online]. Available:
  \url{https://doi.org/10.1007/978-3-030-04630-9_7}
\BIBentrySTDinterwordspacing

\bibitem{anderson2004model}
D.~Anderson and K.~Burnham, ``Model selection and multi-model inference,''
  \emph{Second. NY: Springer-Verlag}, vol.~63, no. 2020, p.~10, 2004.

\bibitem{9149002}
J.~Lin, L.~L. Njilla, and K.~Xiong, ``Robust machine learning against
  adversarial samples at test time,'' in \emph{ICC 2020 - 2020 IEEE
  International Conference on Communications (ICC)}, 2020, pp. 1--6.

\bibitem{DBLP1}
\BIBentryALTinterwordspacing
Y.~Wang, S.~Aeron, A.~S. Rakin, T.~Koike{-}Akino, and P.~Moulin, ``Robust
  machine learning via privacy/rate-distortion theory,'' \emph{CoRR}, vol.
  abs/2007.11693, 2020. [Online]. Available:
  \url{https://arxiv.org/abs/2007.11693}
\BIBentrySTDinterwordspacing

\bibitem{DBLP2}
\BIBentryALTinterwordspacing
H.~Xu and S.~Mannor, ``Robustness and generalization,'' \emph{CoRR}, vol.
  abs/1005.2243, 2010. [Online]. Available:
  \url{http://arxiv.org/abs/1005.2243}
\BIBentrySTDinterwordspacing

\bibitem{DBLP3}
\BIBentryALTinterwordspacing
M.~Shafique, M.~Naseer, T.~Theocharides, C.~Kyrkou, O.~Mutlu, L.~Orosa, and
  J.~Choi, ``Robust machine learning systems: Challenges, current trends,
  perspectives, and the road ahead,'' \emph{CoRR}, vol. abs/2101.02559, 2021.
  [Online]. Available: \url{https://arxiv.org/abs/2101.02559}
\BIBentrySTDinterwordspacing

\bibitem{7888195}
K.~R. Varshney, ``Engineering safety in machine learning,'' in \emph{2016
  Information Theory and Applications Workshop (ITA)}, 2016, pp. 1--5.

\bibitem{doi:10.1177/1687814016668140}
\BIBentryALTinterwordspacing
C.~Park and N.~H. Kim, ``Safety envelope for load tolerance of structural
  element design based on multi-stage testing,'' \emph{Advances in Mechanical
  Engineering}, vol.~8, no.~9, p. 1687814016668140, 2016. [Online]. Available:
  \url{https://doi.org/10.1177/1687814016668140}
\BIBentrySTDinterwordspacing

\bibitem{JMLR:v9:shafer08a}
\BIBentryALTinterwordspacing
G.~Shafer and V.~Vovk, ``A tutorial on conformal prediction,'' \emph{Journal of
  Machine Learning Research}, vol.~9, no.~12, pp. 371--421, 2008. [Online].
  Available: \url{http://jmlr.org/papers/v9/shafer08a.html}
\BIBentrySTDinterwordspacing

\bibitem{MirasierraUQ}
V.~Mirasierra, M.~Mammarella, F.~Dabbene, and T.~Alamo, ``Prediction error
  quantification through probabilistic scaling,'' \emph{IEEE Control Systems
  Letters}, vol.~6, pp. 1118--1123, 2022.

\bibitem{10.5555/2207809}
D.~Barber, \emph{Bayesian Reasoning and Machine Learning}.\hskip 1em plus 0.5em
  minus 0.4em\relax USA: Cambridge University Press, 2012.

\bibitem{10.2307/20445230}
\BIBentryALTinterwordspacing
R.~Herbei and M.~H. Wegkamp, ``Classification with reject option,'' \emph{The
  Canadian Journal of Statistics / La Revue Canadienne de Statistique},
  vol.~34, no.~4, pp. 709--721, 2006. [Online]. Available:
  \url{http://www.jstor.org/stable/20445230}
\BIBentrySTDinterwordspacing

\bibitem{10.5555/1462129}
J.~Quionero-Candela, M.~Sugiyama, A.~Schwaighofer, and N.~D. Lawrence,
  \emph{Dataset Shift in Machine Learning}.\hskip 1em plus 0.5em minus
  0.4em\relax The MIT Press, 2009.

\bibitem{Panchenko}
V.~Koltchinskii and D.~Panchenko, ``Empirical margin distributions and bounding
  the generalization error of combined classifiers,'' \emph{Annals of
  Statistics}, vol.~30, 04 2000.

\bibitem{Evgeniou}
T.~Evgeniou, M.~Pontil, and T.~A. Poggio, ``Regularization networks and support
  vector machines,'' \emph{Advances in Computational Mathematics}, vol.~13, pp.
  1--50, 2000.

\bibitem{price}
D.~Bertsimas and M.~Sim, ``The price of robustness,'' \emph{Operations
  Research}, vol.~52, pp. 35--53, 02 2004.

\bibitem{BENTAL19991}
\BIBentryALTinterwordspacing
A.~Ben-Tal and A.~Nemirovski, ``Robust solutions of uncertain linear
  programs,'' \emph{Operations Research Letters}, vol.~25, no.~1, pp. 1--13,
  1999. [Online]. Available:
  \url{https://www.sciencedirect.com/science/article/pii/S0167637799000164}
\BIBentrySTDinterwordspacing

\bibitem{9408661}
L.~Wang, M.~Han, X.~Li, N.~Zhang, and H.~Cheng, ``Review of classification
  methods on unbalanced data sets,'' \emph{IEEE Access}, vol.~9, pp.
  64\,606--64\,628, 2021.

\bibitem{carlevaro}
A.~Carlevaro and M.~Mongelli, ``Reliable {AI} trough {SVDD} and rule
  extraction,'' \emph{International IFIP Cross Domain (CD) Conference for
  Machine Learning \& Knowledge Extraction (MAKE), CD-MAKE 2021.}, 2021.

\bibitem{SaraIS}
S.~Narteni, V.~Orani, I.~Vaccari, E.~Cambiaso, and M.~Mongelli, ``Sensitivity
  of logic learning machine for reliability in safety-critical systems,''
  \emph{IEEE Intelligent Systems}, pp. 1--1, 2022.

\bibitem{NEURIPS2020_b90c4696}
\BIBentryALTinterwordspacing
J.~Bitterwolf, A.~Meinke, and M.~Hein, ``Certifiably adversarially robust
  detection of out-of-distribution data,'' in \emph{Advances in Neural
  Information Processing Systems}, H.~Larochelle, M.~Ranzato, R.~Hadsell,
  M.~Balcan, and H.~Lin, Eds., vol.~33.\hskip 1em plus 0.5em minus 0.4em\relax
  Curran Associates, Inc., 2020, pp. 16\,085--16\,095. [Online]. Available:
  \url{https://proceedings.neurips.cc/paper/2020/file/b90c46963248e6d7aab1e0f429743ca0-Paper.pdf}
\BIBentrySTDinterwordspacing

\bibitem{9321372}
I.~Stepin, J.~M. Alonso, A.~Catala, and M.~Pereira-Fariña, ``A survey of
  contrastive and counterfactual explanation generation methods for explainable
  artificial intelligence,'' \emph{IEEE Access}, vol.~9, pp. 11\,974--12\,001,
  2021.

\bibitem{9787552}
A.~Carlevaro, M.~Lenatti, A.~Paglialonga, and M.~Mongelli, ``Counterfactual
  building and evaluation via explainable support vector data description,''
  \emph{IEEE Access}, vol.~10, pp. 60\,849--60\,861, 2022.

\bibitem{H_llermeier_2021}
\BIBentryALTinterwordspacing
E.~Hüllermeier and W.~Waegeman, ``Aleatoric and epistemic uncertainty in
  machine learning: an introduction to concepts and methods,'' \emph{Machine
  Learning}, vol. 110, no.~3, pp. 457--506, mar 2021. [Online]. Available:
  \url{https://doi.org/10.1007%2Fs10994-021-05946-3}
\BIBentrySTDinterwordspacing

\bibitem{Alamo:19:SafeApproximations}
T.~Alamo, V.~Mirasierra, F.~Dabbene, and M.~Lorenzen, ``Safe approximations of
  chance constrained sets by probabilistic scaling,'' in \emph{2019 18th
  European Control Conference (ECC)}, 2019, pp. 1380--1385.

\bibitem{Hofmann_2008}
\BIBentryALTinterwordspacing
T.~Hofmann, B.~Schölkopf, and A.~J. Smola, ``Kernel methods in machine
  learning,'' \emph{The Annals of Statistics}, vol.~36, no.~3, jun 2008.
  [Online]. Available: \url{https://doi.org/10.1214%2F009053607000000677}
\BIBentrySTDinterwordspacing

\bibitem{koenker_2005}
R.~Koenker, \emph{Quantile Regression}, ser. Econometric Society
  Monographs.\hskip 1em plus 0.5em minus 0.4em\relax Cambridge University
  Press, 2005.

\bibitem{10.1023/A:1022627411411}
\BIBentryALTinterwordspacing
C.~Cortes and V.~Vapnik, ``Support-vector networks,'' \emph{Mach. Learn.},
  vol.~20, no.~3, p. 273–297, sep 1995. [Online]. Available:
  \url{https://doi.org/10.1023/A:1022627411411}
\BIBentrySTDinterwordspacing

\bibitem{SVDD}
\BIBentryALTinterwordspacing
D.~M.~J. Tax and R.~P.~W. Duin, ``Support vector data description,''
  \emph{Mach. Learn.}, vol.~54, no.~1, p. 45–66, jan 2004. [Online].
  Available: \url{https://doi.org/10.1023/B:MACH.0000008084.60811.49}
\BIBentrySTDinterwordspacing

\bibitem{7056505}
D.~Jia, K.~Lu, J.~Wang, X.~Zhang, and X.~Shen, ``A survey on platoon-based
  vehicular cyber-physical systems,'' \emph{IEEE Communications Surveys \&
  Tutorials}, vol.~18, no.~1, pp. 263--284, 2016.

\bibitem{Xu}
L.~Xu, L.~Y. Wang, G.~Yin, and H.~Zhang, ``Communication information structures
  and contents for enhanced safety of highway vehicle platoons,'' \emph{IEEE
  Transactions on Vehicular Technology}, vol.~63, no.~9, pp. 4206--4220, 2014.

\bibitem{Santini}
S.~Santini, A.~Salvi, A.~S. Valente, A.~Pescapé, M.~Segata, and R.~Lo~Cigno,
  ``A consensus-based approach for platooning with intervehicular
  communications and its validation in realistic scenarios,'' \emph{IEEE
  Transactions on Vehicular Technology}, vol.~66, no.~3, pp. 1985--1999, 2017.

\bibitem{9594676}
A.~Carlevaro and M.~Mongelli, ``A new {SVDD} approach to reliable and
  e{X}plainable {AI},'' \emph{IEEE Intelligent Systems}, vol. in press, pp.
  1--1, 2021.

\bibitem{gentleIntro}
\BIBentryALTinterwordspacing
A.~N. Angelopoulos and S.~Bates, ``A gentle introduction to conformal
  prediction and distribution-free uncertainty quantification,'' 2021.
  [Online]. Available: \url{https://arxiv.org/abs/2107.07511}
\BIBentrySTDinterwordspacing

\bibitem{tutorial_cP}
\BIBentryALTinterwordspacing
G.~Shafer and V.~Vovk, ``A tutorial on conformal prediction,'' 2007. [Online].
  Available: \url{https://arxiv.org/abs/0706.3188}
\BIBentrySTDinterwordspacing

\bibitem{Survey_cP}
\BIBentryALTinterwordspacing
G.~Zeni, M.~Fontana, and S.~Vantini, ``Conformal prediction: a unified review
  of theory and new challenges,'' \emph{CoRR}, vol. abs/2005.07972, 2020.
  [Online]. Available: \url{https://arxiv.org/abs/2005.07972}
\BIBentrySTDinterwordspacing

\end{thebibliography}



\begin{IEEEbiography}[{\includegraphics[width=1in,height=1.25in,clip,keepaspectratio]{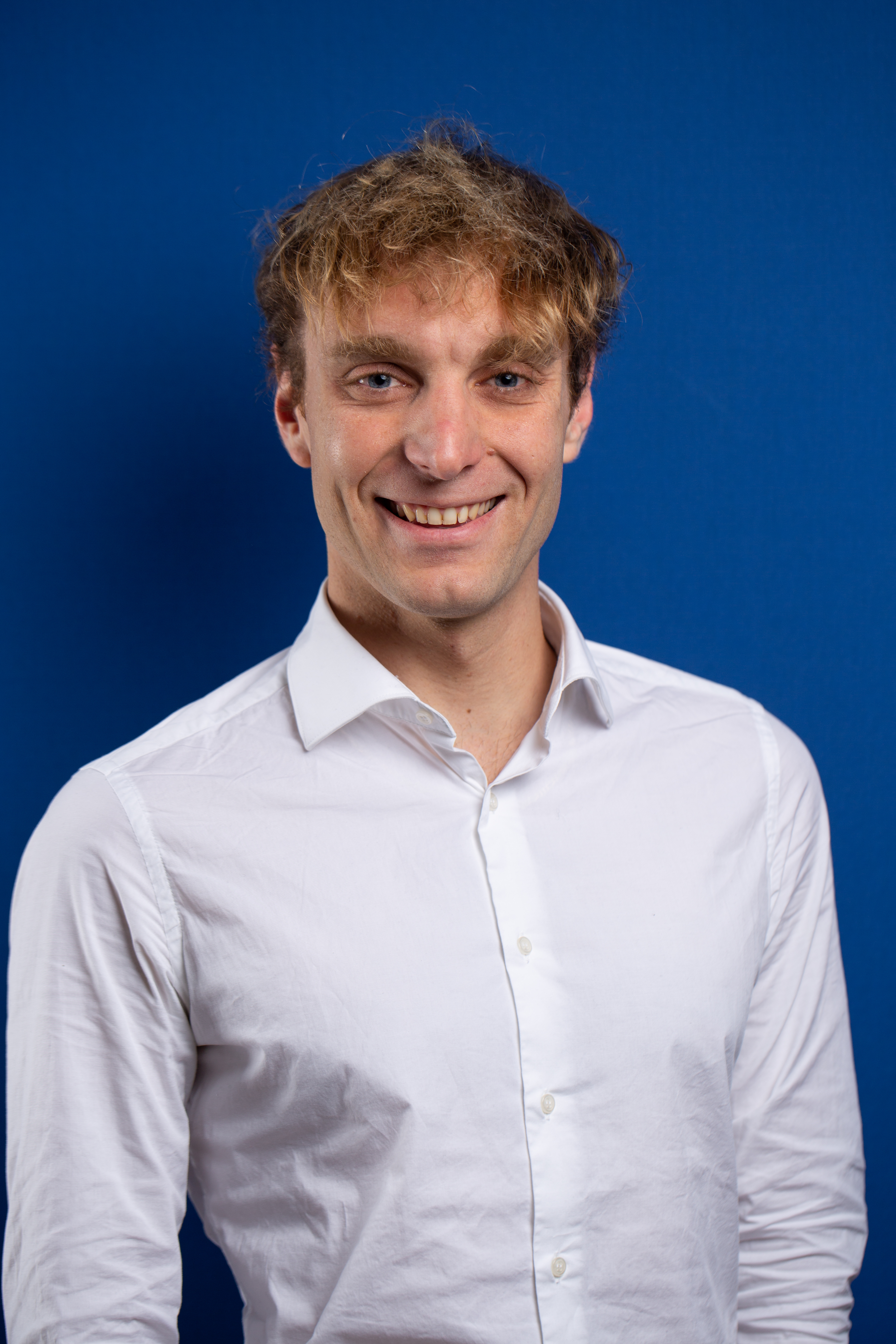}}]{Alberto Carlevaro} {\space} (Student Member, IEEE) received the Master Degree in Applied Mathematics from the University of Genoa, Italy, in 2020 with a physics-mathematics thesis.
He is now a PhD student in electronics at the Department of Electrical, Electronic and Telecommunications Engineering and Naval Architecture (DITEN) of the University of Genoa, in collaboration with CNR-IEIIT and Aitek S.p.A., and now visiting research scholar at EECS department of UC Berkeley. His current fields of research are Machine Learning, Statistical Learning, Explainable AI and Physics Informed Machine Learning. 
\end{IEEEbiography}

\begin{IEEEbiography}[{\includegraphics[width=1in,height=1.25in,clip,keepaspectratio]{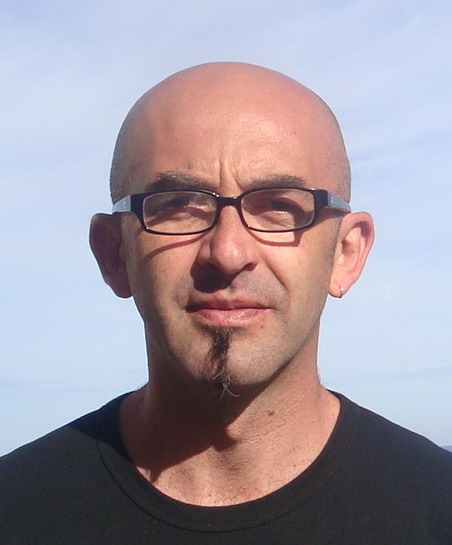}}]{Fabrizio Dabbene}
(Senior Member, IEEE) received the Laurea and Ph.D. degrees from the Politecnico di Torino, Italy, in 1995 and 1999, respectively. He is currently the Director of Research with the Institute IEIIT, National Research Council of Italy (CNR), Milan, Italy, where he is coordinates the Information and Systems Engineering Group. He has held visiting and research positions with The University of Iowa, Penn State University, and the Russian Academy of Sciences, Institute of Control Science, Moscow, Russia. He has authored or coauthored more than 100 research papers and two books. Dr. Dabbene was an Elected Member of the Board of Governors, from 2014 to 2016. He has served as the vice president for publications, from 2015 to 2016. He is currently chairing the IEEE-CSS Italy Chapter. He has also served as an Associate Editor for Automatica, from 2008 to 2014, and IEEE Transactions on Automatic Control, from 2008 to 2012. He is also a Senior Editor of the IEEE Control Systems Society Letters.
\end{IEEEbiography}

\begin{IEEEbiography}[{\includegraphics[width=1in,height=1.25in,clip,keepaspectratio]{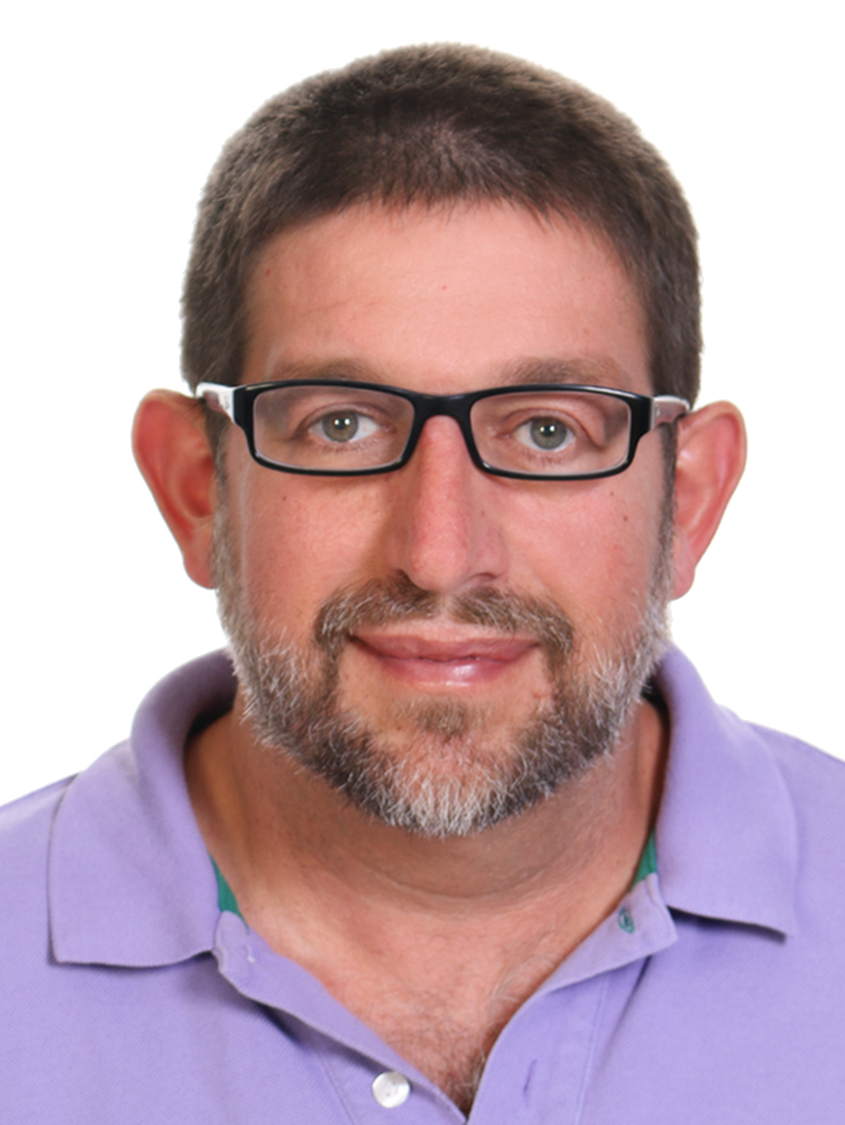}}]{Teodoro Alamo} (Member, IEEE) was born in Spain in 1968. He received the M.Eng. degree in telecommunications engineering from the Polytechnic University of Madrid, Spain, in 1993 and the Ph.D. degree in telecommunications engineering from the University of Seville, Spain, in 1998. From 1993 to 2000, he was an Assistant Professor with the Department of System Engineering and Automatic Control, University of Seville, where he was an Associate Professor from 2001 to 2010 and has been a Full Professor since March 2010. He was at the Ecole Nationale Superieure des T\'el\'ecommunications (Telecom Paris) from September 1991 to May 1993. Part of his Ph.D. was done at RWTH Aachen, Alemania, from June to September 1995. He is the author or coauthor of more than 200 publications including books, book chapters, journal papers, conference proceedings, and educational books. (google scholar profile available at http://scholar.google.es/citations?user=W3ZDTkIAAAAJ\&hl=en). His current research interests include decision making, model predictive control, data-driven methods, randomized algorithms, and optimization strategies.
\end{IEEEbiography}

\begin{IEEEbiography}[{\includegraphics[width=1in,height=1.215in,clip,keepaspectratio]{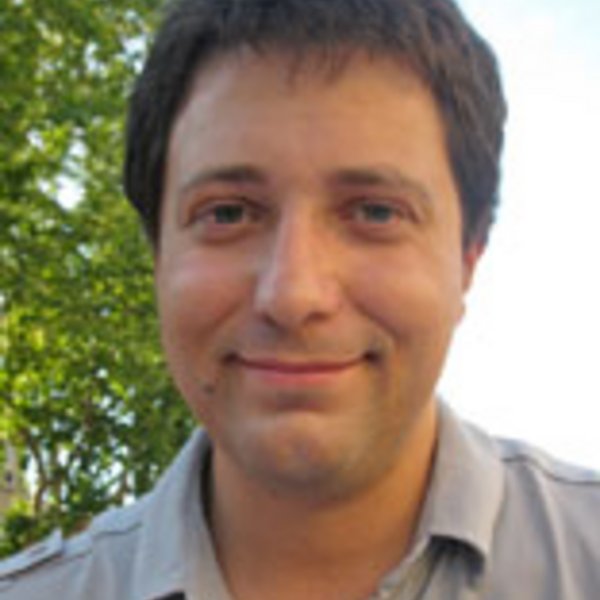}}]{Maurizio Mongelli} (Member, IEEE)
obtained his PhD. Degree in Electronics and Computer Engineering from the University of Genoa in 2004. 
He worked for Selex and the Italian Telecommunications Consortium (CNIT) from 2001 until 2010. 
He is now a researcher at CNR-IEIIT, where he deals with machine learning applied to health and cyber-physical systems. He is co-author of over 100 international scientific papers, 2 patents and is participating in the SAE G-34/EUROCAE WG-114 AI in Aviation Committee.
\end{IEEEbiography}

\end{document}